\documentclass[lettersize,journal]{IEEEtran}
\usepackage{amsmath,amsfonts}
\usepackage{algorithmic}
\usepackage{algorithm}
\usepackage{array}
\usepackage[caption=false,font=normalsize,labelfont=sf,textfont=sf]{subfig}
\usepackage{textcomp}
\usepackage{stfloats}
\usepackage{url}
\usepackage{verbatim}
\usepackage{graphicx}
\usepackage{cite}

\usepackage{booktabs}       
\usepackage{nicefrac}       
\usepackage{microtype}      
\usepackage[table]{xcolor}         
\usepackage[pdftex]{hyperref}
\hypersetup{
	colorlinks=true,
	linkcolor=purple,
	filecolor=purple,
	urlcolor=purple,
	citecolor=cyan,
}
\usepackage{multirow}
\usepackage{colortbl} 
\definecolor{tabletitle}{gray}{.8}
\definecolor{ours}{gray}{.95}
\definecolor{ggray}{RGB}{127,127,127}
\definecolor{reda}{RGB}{202,0,0}
\definecolor{redb}{RGB}{217,148,143}
\definecolor{myyellow}{RGB}{190,144,0}
\definecolor{mygreen}{RGB}{0,136,51}
\definecolor{myblue}{RGB}{0,102,204}
\newcolumntype{B}{!{\vrule width 1pt}}
\usepackage{pifont}
\newcommand{\blank}{—}

\begin{document}

\title{CAVER: Cross-Modal View-Mixed Transformer for Bi-Modal Salient Object Detection}

\author{
  Youwei Pang, Xiaoqi Zhao, Lihe Zhang and Huchuan Lu
  \thanks{Y. Pang, X. Zhao, L. Zhang and H. Lu are with the School of Information and Communication Engineering, Dalian University of Technology, Dalian, China (e-mail: lartpang@mail.dlut.edu.cn; zxq@mail.dlut.edu.cn; zhanglihe@dlut.edu.cn; lhchuan@dlut.edu.cn). This work was supported by the National Natural Science Foundation of China \#62276046 and the Liaoning Natural Science Foundation \#2021-KF-12-10.}
  \\
  \url{https://github.com/lartpang/CAVER}
}

\markboth{Journal of \LaTeX\ Class Files,~Vol.~14, No.~8, August~2021}%
{Shell \MakeLowercase{\textit{et al.}}: A Sample Article Using IEEEtran.cls for IEEE Journals}


\maketitle

\begin{abstract}
  Most of the existing bi-modal (RGB-D and RGB-T) salient object detection methods utilize the convolution operation and construct complex interweave fusion structures to achieve cross-modal information integration.
  The inherent local connectivity of the convolution operation constrains the performance of the convolution-based methods to a ceiling.
  In this work, we rethink these tasks from the perspective of global information alignment and transformation.
  Specifically, the proposed \underline{c}ross-mod\underline{a}l \underline{v}iew-mixed transform\underline{er} (CAVER) cascades several cross-modal integration units to construct a top-down transformer-based information propagation path.
  CAVER treats the multi-scale and multi-modal feature integration as a sequence-to-sequence context propagation and update process built on a novel view-mixed attention mechanism.
  Besides, considering the quadratic complexity w.r.t. the number of input tokens, we design a parameter-free patch-wise token re-embedding strategy to simplify operations.
  Extensive experimental results on RGB-D and RGB-T SOD datasets demonstrate that such a simple two-stream encoder-decoder framework can surpass recent state-of-the-art methods when it is equipped with the proposed components.
  Code and pretrained models will be available at \href{https://github.com/lartpang/CAVER}{the link}.
\end{abstract}


\section{Introduction}

\IEEEPARstart{S}{alient} object detection (SOD) aims to identify the most significant objects or regions in images or videos from various visual scenes.
It plays a fundamental and important role in many computer vision tasks, such as
semantic segmentation~\cite{WSSS-STC},
medical image segmentation~\cite{MIS-MSNet,PraNet}
video object segmentation~\cite{USVOS-COSNet,VSOD-ShiftingFDP,MS-APS},
person re-identification~\cite{PersonReID-USL},
camouflaged object detection~\cite{COD10K,ConcealedObjectDetection,ZoomNet-CVPR2022},
image editing~\cite{ImageEditing}
and compression~\cite{Compression}.

Although many purely CNN-based methods~\cite{F3Net,LDF,GateNet,DFI,MINet} achieve quite promising results on the RGB SOD task, they still struggle with challenges of complex or low-contrast scenes, and obscured or indistinguishable objects.
To this end, some recent studies have attempted to introduce additional information, such as depth~\cite{RGBDSODSurveyZhou}, light field~\cite{LFSODSurveyFu} and thermal infrared~\cite{VT5000-ADF} images, to make the model comfortable with complicated and diverse natural scenes.
The depth map can explicitly provide complementary spatial structure information and the relative position relationship between objects,
and the infrared map based on the thermal radiation can better present the overall shape of the object in various complex scenarios.
They help to distinguish different objects and alleviate the above issues.
Therefore, recent research is gradually shifting towards the integration of modality-specific and modality-complementary cues from RGB and depth/thermal images to excavate and capture objects of interest.
At the same time, long-range context and contrast information play a key role in identifying and locating salient objects.
However, the purely convolution-based architectures \footnote{Their core components are the convolution, which is different from our transformer-based structure.} possibly encounter the performance bottleneck, which is caused by the localized convolution operation and the fixedness of the learned parameters.
Hence, the introduction of a new architecture becomes more and more important.

\begin{figure}[t]
  \centering
  \includegraphics[width=0.8\linewidth]{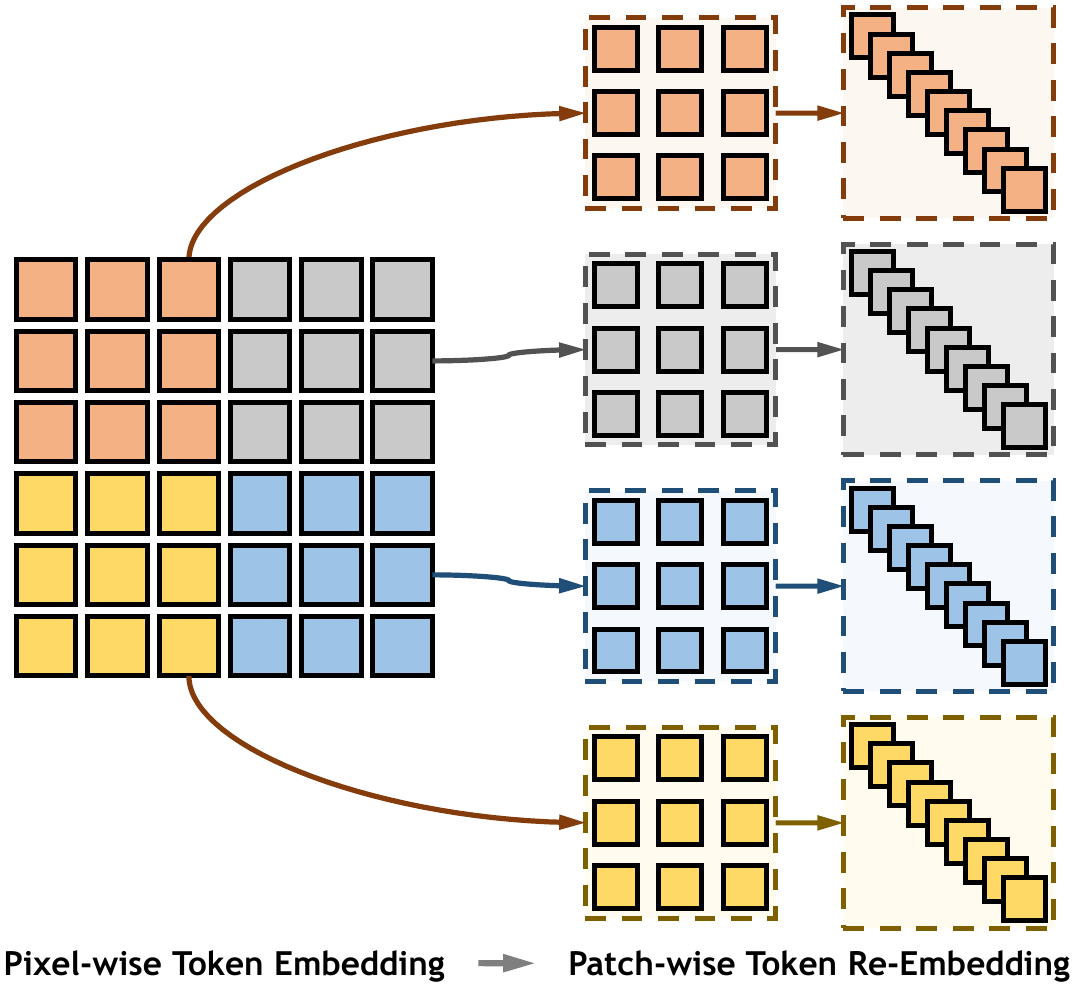}
  \caption{
    Patch-wise token re-embedding (PTRE).
    Before matrix multiplication, the parameter-free PTRE is used to reshape features.
    Thus, pixel-wise tokens are aggregated and converted into patch-wise tokens.
  }
  \label{fig:patchembedding}
\end{figure}

The transformer~\cite{Transformer} has superior performance against existing state-of-the-art CNNs in several computer vision tasks~\cite{ViT,Swin}, which can be attributed to its powerful ability to extract features and model long-range dependencies.
In this paper, we introduce the transformer to design a simple yet effective RGB-D SOD model, CAVER.
A novel transformer-based top-down multi-level structure is engineered for the cross-modal fusion, which can be easily assembled to a CNN feature extractor, like ResNets~\cite{Resnet,Res2Net,ResNet-D}.
We leverage the self- and cross-attention to learn feature alignment where each element gathers information from other elements in the global sequence based on semantic similarity.
This can naturally achieve the transformation and enhancement of intra-modal features and the matching and fusion of inter-modal features.
However, a prominent problem of the transformer is the unbearable quadratic computational complexity caused by the attention operation based on all elements of the feature map.
To cope with this, we introduce the patch embedding into the attention block.
Instead of using the patch embedding to extract and compress image features in the input stage~\cite{ViT} or at the beginning of each stage corresponding to different feature scales~\cite{Swin}, we apply it to improve each matrix multiplication of self- and cross-attention with the pixel-wise tokens as the input.
We call this parameter-free operation ``patch-wise token re-embedding'' (PTRE) as shown in Fig.~\ref{fig:patchembedding}.
When learning multi-scale and high-resolution features, such a design can reduce computation and memory costs by a factor of $p^{2}$ and $p^{4}$ ($p$ is the side length of the patch), respectively.
At the same time, considering that the existing transformer-based methods pay too much attention to the spatial context information in the process of the feature alignment while ignoring the positive role of the channel context.
Therefore we introduce a parallel channel attention branch into the attention operation and propose the view-mixed attention (VMA) block.
Global context and object detail information have positive value for the SOD task, so the convolutional design is introduced into the feed forward network that follows the attention operation to enhance the perception of local details.
As a result, the proposed components can work together to effectively explore and excavate global and local cues in the feature decoding process.

Our contributions can be summarized as:
\begin{itemize}
  \item We introduce the transformer to rethink the bi-modal SOD modeling from a sequence-to-sequence perspective, which gains better interpretability.
  \item We build a top-down transformer-based information propagation path enhanced by the view-mixed attention block, which can align the features of RGB and depth/thermal modalities and fully exploit the inter- and intra-modal information from spatial and channel views.
  \item We boost the matrix operation in the attention by using the patch-wise token re-embedding, which improves the efficiency of the transformer for multi-scale and high-resolution features. And aided by the convolutional feed forward network, the locality of features can be further enhanced, and key cues in both global and local contexts can be fully perceived and explored.
  \item Extensive experiments demonstrate that the proposed model outperforms recent methods on seven RGB-D SOD datasets and three RGB-T SOD datasets.
\end{itemize}

\section{Related Work}

\paragraph{Visual Attention.}
Humans can quickly capture significant objects or regions in a scene.
The modeling and investigation of such an ability is a fundamental and critical problem in computer vision, i.e. visual attention mechanism.
Research in this area can be divided into two different directions: one is to explore where the observer is looking, i.e. eye fixation prediction~\cite{SOD-ITTI}, and the other is to locate and segment completely visually attractive regions, i.e. salient object detection (SOD)~\cite{SOD-Segmentation,SOD-FrequencyTuned}.
In this paper, we focus on the latter.
With the UNet-like~\cite{UNet} architecture becoming the dominant paradigm for SOD, the proliferation of CNN-based methods has driven the rapid development of SOD in recent years~\cite{CMMSODSurvey,WWGSODSurvey}.

\paragraph{Bi-modal Salient Object Detection.}
In the bi-modal SOD field, RGB-D and RGB-T are two thriving and important branches.
They additionally introduce depth or thermal infrared information, respectively, which can provide a more comprehensive understanding of the scene.
Roughly speaking, the existing methods can be broadly classified into three categories depending on the cross-modal fusion strategy: early fusion, intermediate fusion, and late fusion.
\textbf{Early fusion} methods directly combine the low-level information of two modalities, e.g., concatenating the inputs~\cite{NLPR,DANet,SIP} or integrating their low-level features~\cite{RGBDSOD-DeepFusion}.
Although such a scheme may reduce the number of model parameters, this also makes it difficult to control the interference of noise within different modalities to the whole model.
Unlike early fusion, \textbf{late fusion} methods generally adopt a dual-stream structure and focus more on the cross-modal fusion of high-level features~\cite{CTMF,HDFNet,DisenFuse} or final predictions~\cite{AFNetRGBD}.
The semantic information enriched by high-level features has a positive guide for good prediction results.
In fact, both low-level and high-level features have the same importance in the SOD task, and they complement each other.
The low-level features can provide rich texture and structural scene perception, which is missing in the high-level features.
Hence, an \textbf{intermediate fusion} strategy that utilizes both simultaneously is gradually becoming the mainstream of bi-modal SOD methods~\cite{TANet,MMCI,CPFP,DUTRGBD,CMWNet,CoNet,SSLSOD-RGBDSOD,MMFT-RGBDSOD,BiANet-RGBDSOD-journal,JLDCF-RGBDSOD-journal,HAINet-RGBDSOD,CCAFNet-RGBDSOD,RD3D-RGBDSOD,ECFFNet-RGBTSOD,CGFNet-RGBTSOD,CSRNet-RGBTSOD,MIDD-RGBTSOD}.
Our proposed method can also be classified into this category.
These algorithms usually construct various cross-modal interaction strategies based on the CNN structure, which typically draw on the plug-and-play modules or their variants to enhance and rectify the representation of features such as ASPP~\cite{Deeplab} or DenseASPP~\cite{DenseASPP}, PPM~\cite{PPM}, convolutional channel/spatial attention blocks~\cite{SENet,CBAM,SKNet} in~\cite{DUTRGBD,CoNet,DANet,S2MA,UCNet-RGBDSOD-journal,COME15K-CMINet,BBSNet-RGBDSOD-journal,JLDCF-RGBDSOD-journal,ECFFNet-RGBTSOD,CGFNet-RGBTSOD,CSRNet-RGBTSOD,MIDD-RGBTSOD} and ConvLSTM~\cite{ConvLSTM} in~\cite{DUTRGBD}.
Specifically, a novel joint learning and densely cooperative fusion architecture through a siamese network are designed in~\cite{JLDCF-RGBDSOD-journal}.
In~\cite{BBSNet-RGBDSOD-journal}, the bifurcated backbone strategy and depth-enhanced module are proposed to excavate informative cues and fuse the two modalities in a complementary way.
And~\cite{SPNet-RGBDSOD-journal} explores both the shared information and modality-specific properties.
Uncertainty-aware stochastic framework~\cite{UCNet-RGBDSOD-conference,UCNet-RGBDSOD-journal} and mutual information minimization regularization~\cite{COME15K-CMINet} are also introduced to optimize the interaction process of two modalities.
Unlike them, we introduce a global sequence perspective for feature enhancement and interaction.
This can effectively fill the lack of contextual information caused by the local reception field of the convolution operation.

\begin{figure*}[!t]
  \centering
  \includegraphics[width=0.9\linewidth]{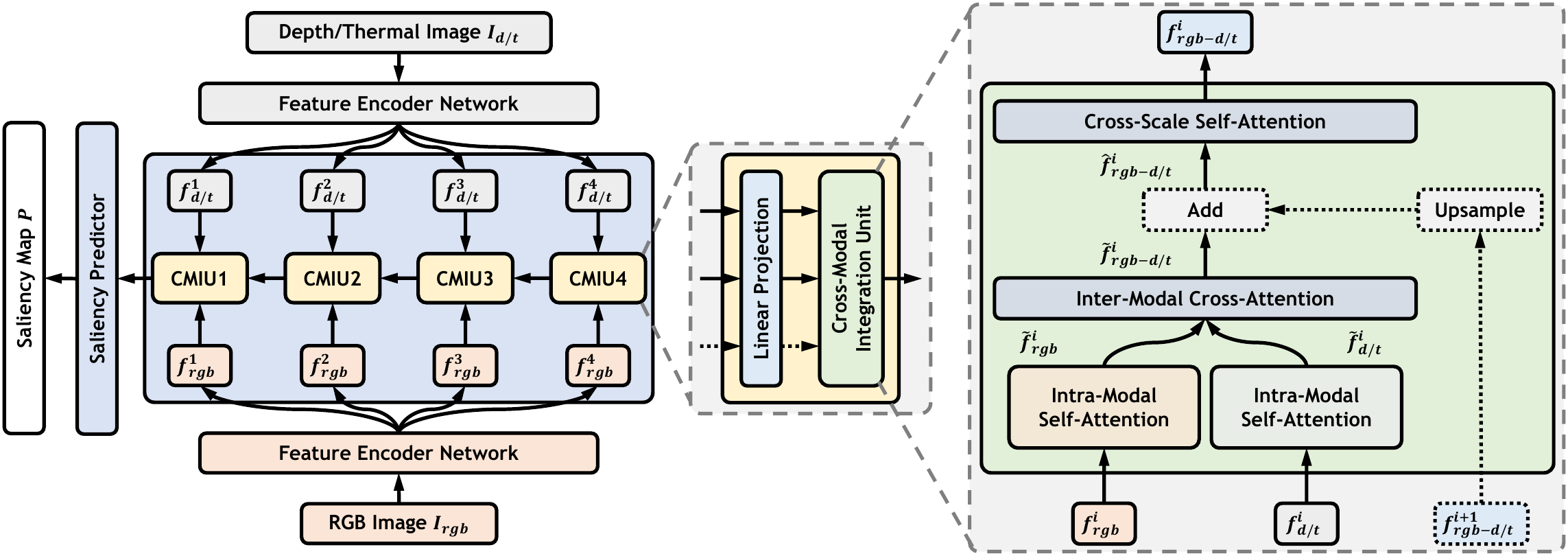}
  \caption{
    The overview of the proposed model.
    This is a dual-stream encoder-decoder architecture with a very simple and straightforward form.
    Note that the dashed line denotes an optional path for the decoder.
    In our model, the CMIU4 only contains two inputs $f^{4}_{rgb}$ and $f^{4}_{d/t}$ and $\hat{f}^{4}_{rgb-d/t}=\tilde{f}^{4}_{rgb-d/t}$.
    The feature $f^{i+1}_{rgb-d/t}$ exists in CMIU1-3, which is upsampled using bilinear interpolation in the 2D form.
  }
  \label{fig:net}
\end{figure*}

\paragraph{Attention-Based Model.}
On the one hand, as mentioned earlier, the attention mechanism is closely related to the SOD task.
Some methods~\cite{DUTRGBD,UCNet-RGBDSOD-conference,UCNet-RGBDSOD-journal,CoNet} apply convolutional channel and spatial attention blocks~\cite{SENet,CBAM,SKNet}.
And recent methods design some task-friendly variants.
Inspired by the dynamic convolution~\cite{SpatialAttentionStudy-DJF},~\cite{HDFNet} design the dynamic dilated pyramid module with position-specific and image-specific multi-scale filters to provide cross-modal contextual guidance for the RGB feature.
A saliency-guided bilateral attention module in~\cite{BiANet-RGBDSOD-journal} is proposed to capture meaningful foreground and background complementary information.
Although these data-adaptive feature enhancement methods in spatial or channel form can improve the flexibility and expressiveness of the model, these convolution-based strategies do not model long-range dependencies well and still have a large room for improvement.
The most related works~\cite{S2MA,TriTransNet-RGBDSOD} introduce the non-local block~\cite{NonLocalNet} or self-attention block~\cite{Transformer}, but they are only used to enhance the high-level feature interaction in the spatial view and the model body is still limited by the CNN architecture.
Besides, limited by the large computational cost of the original self-attention operation, they give up the positive gain that contextual information can bring in the processing and fusion of shallow bi-modal features.
In contrast, our channel and spatial view-mixed approach achieves further exploration by thoroughly building the multi-level cross-modal fusion scheme from a sequence-to-sequence transformation perspective.
On the other hand, the transformer~\cite{Transformer} is built on the similarity-based attention mechanism, which has shown powerful performance in natural language processing and is receiving more and more interest from researchers in computer vision.
The recent remarkable performance of vision transformers~\cite{ViT,Swin} reflects that the transformer is a general and effective architecture to transform features.
Most of them use the transformer to extract image features for the classification task.
SETR~\cite{SETR} is a pioneering work of applying the transformer to the segmentation field and it uses ViT~\cite{ViT} to extract features and builds a lightweight convolutional decoder to obtain predictions.
Similarly to it, the existing methods tend to explore the application of the transformer in the encoder, and there is little work focusing on the decoder.
In the segmentation task, the decoder plays an important role and both transformer-based encoder and decoder are worthy of being explored.
So different from the existing methods which propose the transformer-based encoder, our work is more inclined to explore the design of the transformer-based decoder, especially for cross-modal tasks such as RGB-D SOD and RGB-T SOD.
We use it to decode the extracted features from RGB and depth/thermal images and build a simple encoder-decoder architecture for these two bi-modal SOD tasks.
Since these bi-modal tasks require considering two modalities, their integration and alignment are the key points of our model design.
This work and the existing methods can complement each other.
In addition, we apply the transformer to multi-scale high-resolution features, which faces the computational pressure from higher-resolution features.

\begin{figure*}[!t]
  \centering
  \subfloat[]{
    \includegraphics[width=0.24\linewidth]{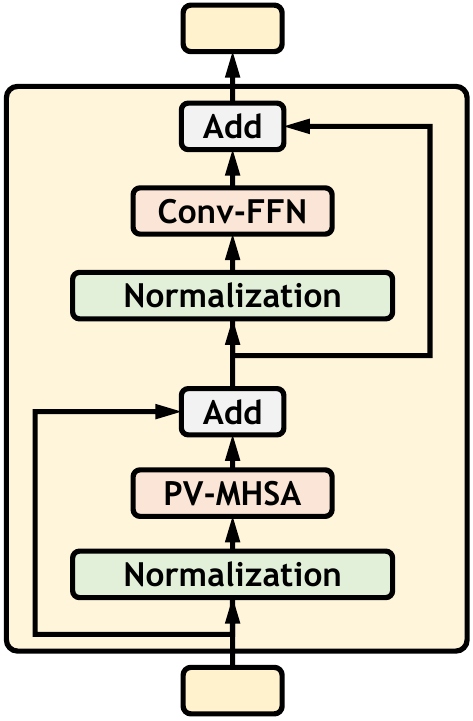}
    \label{fig:sa}
  }%
  \hfil
  \subfloat[]{
    \includegraphics[width=0.34\linewidth]{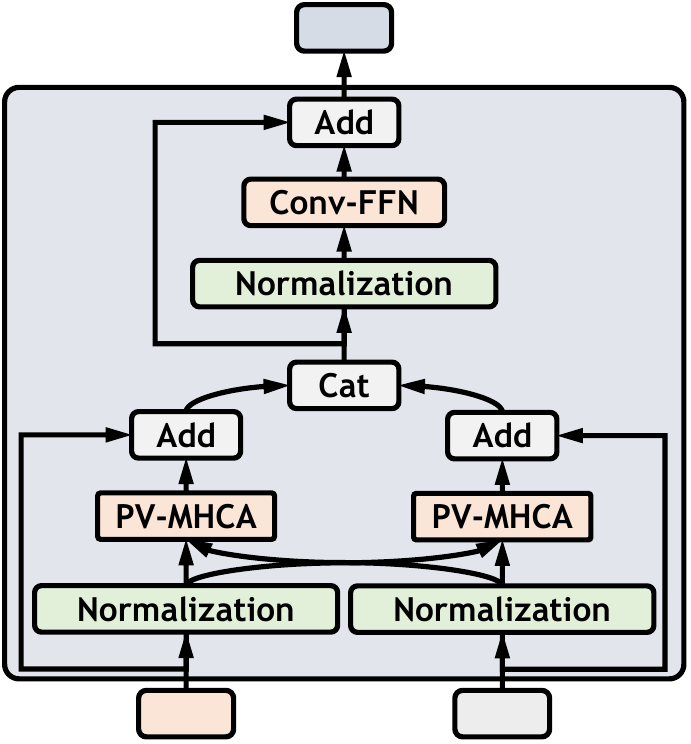}
    \label{fig:ca}
  }
  \hfil
  \subfloat[]{
    \includegraphics[width=0.26\linewidth]{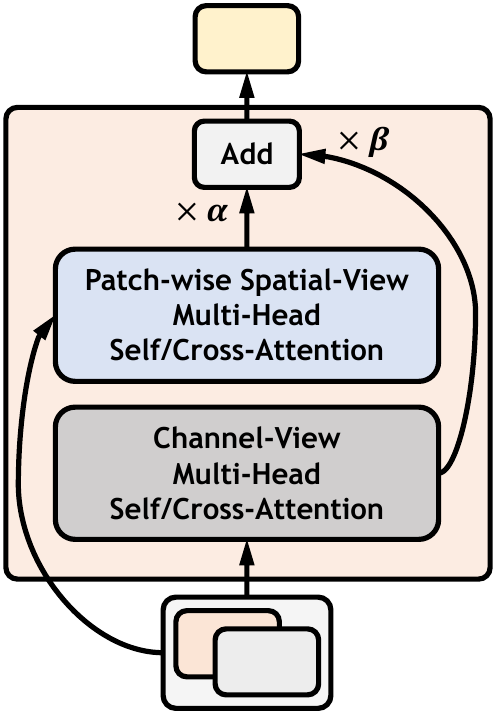}
    \label{fig:ma}
  }
  \caption{
    Patch-wise view-mixed attention blocks of the proposed network.
    ``Normalization'' denotes the BN~\cite{BatchNorm} layer in our method.
    (a) Self-attention block.
    (b) Cross-attention block.
    (c) View-mixed attention operation.
  }
  \label{fig:attention}
\end{figure*}

\begin{figure}[!t]
  \centering
  \includegraphics[width=0.8\linewidth]{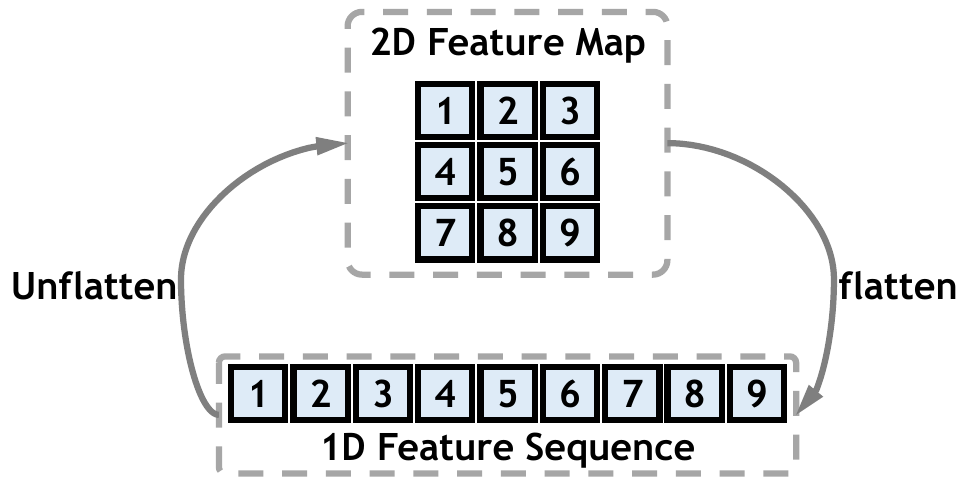}
  \caption{
    Conversion between the 2D feature map and the 1D feature sequence.
  }
  \label{fig:reshape}
\end{figure}

\section{Our Method: CAVER}

In this paper, we propose CAVER, a simple yet effective cross-modal feature integration network based on the transformer.
In the following, we first show the overview of the model and then describe the details of each component.

\subsection{Network Overview}

In our approach, the widely used CNN backbone acts as the feature encoder network for two modalities: RGB and depth/thermal.
As shown in Fig.~\ref{fig:net}, the features with different scales (i.e. $\{f^{i}_{rgb}\}^4_{i=1}$ from the RGB image $I_{rgb}$ and $\{f^{i}_{d/t}\}^4_{i=1}$ from the depth/thermal image $I_{d/t}$) from each intermediate layer of the backbone are fed directly to the corresponding stage in the proposed transformer-based information propagation path.
These multi-scale features abound with the cues about appearances, boundaries, textures, and more complex semantic concepts from the RGB modality, relative spatial location relationships between objects from the depth modality, and the integrity and thermal radiation property of the object self from the thermal infrared modality.
By absorbing shallow features from top to down, this serialized cross-modal information aggregation process gradually recovers the high-resolution representation.
And the salient object regions are enriched and shaped.
Eventually, the refined features are mapped as the final single-channel saliency map $\mathcal{P}$ via a simple saliency predictor which contains a bilinear interpolation operation, some convolution layers, and a sigmoid activation.
It is supervised by the binary ground truth $\mathcal{G}$.

\subsection{Transformer-based Information Propagation Path (TIPP)}\label{sec:tipp}

The TIPP is proposed mainly to process and integrate top-down features from both RGB and depth/thermal modalities with different scales, which consists of four cascaded cross-modal integration units (CMIUs), as shown in Fig.~\ref{fig:net}.
Among them, the $4^{th}$ CMIU is only responsible for the integration of cross-modal information.
It takes feature maps $f^{4}_{rgb}$ and $f^{4}_{d/t}$ with the same scale as input.
While each of the remaining CMIUs needs to additionally integrate the output feature map from the adjacent higher level.
In addition, each input map of the CMIU needs to be converted into a $D$-dimensional pixel-wise embedding map in a separate embedding layer.
Empirically, D is set to 64, which helps generate more compact feature embeddings and reduces memory cost.
Before the integration of modalities through the inter-modal cross attention block, $f_{rgb}$ and $f_{d/t}$ are first reconstructed and self-reinforced by an intra-modal self-attention block, respectively.
For the following cross-scale self-attention block, if there exists the feature $f^{i+1}_{rgb-d/t}$ from the previous CMIU, the input $\hat{f}^{i}_{rgb-d/t}$ is the sum of the upsampled $f^{i+1}_{rgb-d/t}$ and the output $\tilde{f}^{i}_{rgb-d/t}$ of the inter-modal cross-attention block, otherwise, the input is equal to $\tilde{f}^{i}_{rgb-d/t}$.
The resolution of $f^{1}_{rgb-d/t}$ is a quarter of that of the input $I_{rgb}$ and $I_{d/t}$.

\subsection{Intra-Modal/Cross-Scale Self-Attention (IMSA/CSSA)}\label{sec:imcssa}

The IMSA and CSSA blocks play different roles, but their structures are the same.
The specific details of the self-attention block can be found in Fig.~\ref{fig:sa}, which is characterized by a patch-wise view-mixed multi-head self-attention (PV-MHSA), a convolutional feed forward network (Conv-FFN), normalization layers and residual connectors.
The PV-MHSA follows a pre-normalization~\cite{PreAct-He,PreNorm-Xiong} layer and they are wrapped by a residual connector.
The output is then passed to a pre-normalized Conv-FFN whose input and output are similarly connected in a residual way.
The overall process can be expressed as:
\begin{equation}
  \begin{split}
    \label{equ:sablock}
    X & = \text{PV-MHSA}(\text{Norm}(X)) + X, \\
    X & = \text{Conv-FFN}(\text{Norm}(X)) + X,
  \end{split}
\end{equation}
where $X \in \mathbb{R}^{N \times D}$ refers to the flattened image features.
$N=H \times W$ and $D$ denote the number of pixel tokens and the embedding dimension.

\paragraph{Multi-Head Self-Attention (MHSA).}\label{par:mhsa}
At first, we introduce its original form in the transformer~\cite{Transformer}.
This operation is actually a feature alignment process, which computes the feature correlation to reconstruct the query itself.
And a single head of it can be defined as:
\begin{equation}
  \begin{split}
    \label{equ:shsa}
    Y_{h} = \text{Attn}(Q_{h}, K_{h}^{\top}, V_{h}, e)
    & = \text{Softmax}(\frac{Q_{h}K_{h}^{\top}}{e})V_{h}, \\
    [Q_{h}, K_{h}, V_{h}] & = X[\mathbf{W}_{q}, \mathbf{W}_{k}, \mathbf{W}_{v}],
  \end{split}
\end{equation}
where $Q_{h}, K_{h}, V_{h}$ are a single head of query, key and value, respectively.
$\mathbf{W}_{q}, \mathbf{W}_{k}, \mathbf{W}_{v} \in \mathbb{R}^{D \times D/N_{h}}$ are the corresponding projection matrices.
$N_{h}$ is the number of heads and $e=\sqrt{D/N_{h}}$ is a scaling factor.
The outputs from different heads are concatenated together and fused by a dense layer.
The final output $Z \in \mathbb{R}^{N \times D}$ can be expressed as $Z = [Y_{1}, \dots, Y_{N_h}]\mathbf{W}_{o}$ and $\mathbf{W}_{o} \in \mathbb{R}^{D \times D}$ is an output projection layer.
It is noted that the dot product operation in the attention matrix $Q_{h}K_{h}^{\top}$ has a quadratic complexity w.r.t. the input sequence length, i.e. $N^{2}$, which limits it to handling multi-scale high-resolution features.
In addition, the current MHSA only considers feature alignment on the spatial view, while ignoring the potential value of the channel view.
With these considerations in mind, two improvements are made in our method:
1) The patch-wise token re-embedding strategy (PTRE) is designed to reduce the computational complexity, which ensures the ability of the model to integrate a wider range of contextual information.
2) The feature reconstruction in the channel view is also introduced to build the view-mixed attention (VMA) based on MHSA.
Based on the proposed PTRE and VMA, we can construct a lighter PV-MHSA to replace the original MHSA, which has a more efficient calculation process and a better modeling capability.

\paragraph{Patch-wise Token Re-Embedding (PTRE).}\label{par:ptre}
The PTRE is applied to improve the matrix operation from the pixel-wise form to the patch-wise form compared with the MHSA, which reduces the complexity by a factor of $p^2$.
Here, $p^2$ is the number of elements in a patch.
Specifically, $Q_{h}$, $K_{h}$ and $V_{h}$ in Equ.~\ref{equ:shsa} are unflattened to the 2D form (Fig.~\ref{fig:reshape}) and adjusted by the PTRE operation.
In the PTRE, the map with $D/N_{h}$-dimensional embedding is unfolded to $N/p^2 \times Dp^2/N_{h}$ (Fig.~\ref{fig:patchembedding}).
After the operation in Equ.~\ref{equ:shsa}, the patch-wise result is reshaped back to $N \times D/N_{h}$ and subsequent operations are consistent with those of the MHSA.

\paragraph{View-Mixed Attention (VMA).}\label{par:vma}
In our method, the channel-view self-attention is introduced to enhance the form of self-attention shown in Equ.~\ref{equ:shsa} which only gathers the information from the spatial view.
The main difference between the two is that the objects for pairwise similarity calculation are changed from spatial locations in the feature sequence to independent feature channels.
Specifically, the form of the calculation in Equ.~\ref{equ:shsa} is consequently transformed into the following form:
\begin{equation}
  \begin{split}
    \label{equ:vma}
    Y_{h}^{\top} & = \text{Attn}(Q_{h}^{\top}, K_{h}, V_{h}^{\top}, n),
  \end{split}
\end{equation}
where $n=\sqrt{N}$ is a scaling factor.
In the proposed block, the two operations are executed in parallel, and the output features $Z_{s}$ and $Z_{c}$ from the spatial and channel branches are combined by learnable weights $\alpha \in [0, 1]$ and $\beta \in [0, 1]$ to obtain the output of the VMA: $Z=\alpha Z_{s} + \beta Z_{c}$.
Compared with the single spatial token attention in Equ.~\ref{equ:shsa}, the computational cost is reduced from $2N^2D$ to $2N^2D/p^2+2ND^2$, and the memory cost is reduced from $N_hN^2+ND$ to $N_hN^2/p^4+D^2/N_h+2ND$.
In our default setting ($D=64$ and $N_h=2$), when $N>65$ (i.e., the length and width of the feature map are both greater than 8 and it is a very common scenario), the proposed view-mixed attention is more efficient than the standard attention form.

\paragraph{Convolutional Feed Forward Network (Conv-FFN).}\label{par:cffn}
Although the transformer can better model the long-range interaction, the local correlation still has a strong practical value as a kind of inductive bias for image data in the vision task.
The original position-wise feed forward network (FFN) in the transformer only performs the separate channel transformation for each element of the sequence, which lacks attention to the local context.
Hence, we adapt the FFN by using the convolution operation.
The original two fully-connected layers are replaced with the common combination of ``$3 \times 3$ convolution $\rightarrow$ batch normalization $\rightarrow$ activation'' and finally a $1 \times 1$ convolution layer is utilized to obtain the feature with the same dimension as the output of the original FFN.

\subsection{Inter-Modal Cross-Attention (IMCA)}\label{sec:imca}

The IMCA block contains two streams (i.e., RGB and depth/thermal) and the multi-head cross-attention (MHCA) is the protagonist here.
The MHCA is very similar to the MHSA, the only difference is that the sources of information for $Q$ and $K/V$ are no longer the same.
This change in form allows it to be used to construct interactions between different information sources.
In our method, the cross-attention block (Fig.~\ref{fig:ca}) is used to associate and interact information between modalities.
Its input is from the outputs of two separated IMSA blocks.
Similarly to the self-attention block, the input features are also normalized.
Then, they are passed into the patch-wise view-mixed multi-head cross-attention (PV-MHCA) equipped with the PTRE and the VMA for cross-modal alignment and enhancement.
Next, the features $Z^{rgb}$ and $Z^{d/t}$ from two modality streams are concatenated, normalized and then locally enhanced by a cascaded Conv-FFN.
Finally, the output is the sum of the results from the PV-MHCA and the Conv-FFN.
Specifically, the RGB stream can be formulated as:
\begin{equation}
  \begin{split}
    \label{equ:shca}
    Z^{rgb} & = \alpha^{rgb}Z^{rgb}_{s} + \beta^{rgb}Z^{rgb}_{c}, \\
    Z^{rgb}_{s} & = [\text{Attn}(Q^{rgb}_{h}, K^{d/t\top}_{h}, V^{d/t}_{h}, e)]^{N_h}_{h=1}\mathbf{W}_{s}, \\
    Z^{rgb\top}_{c} & = [\text{Attn}(Q^{rgb\top}_{h}, K^{d/t}_{h}, V^{d/t\top}_{h}, n)]^{N_h}_{h=1}\mathbf{W}_{c},
  \end{split}
\end{equation}
where $[\dots]^{N_h}_{h=1}$ represents the concatenation operation for all heads and $\mathbf{W}_{s}$ and $\mathbf{W}_{c}$ are output projection matrices corresponding to different views to fuse these heads, similar to the PV-MHSA.
To get the output $Z^{d/t}$ of the depth/thermal stream, just replace inputs $Q^{rgb}$, $K^{d/t}$ and $V^{d/t}$, and learnable weights $\alpha^{rgb}$ and $\beta^{rgb}$: $Q^{rgb} \rightarrow Q^{d/t}, K^{d/t} \rightarrow K^{rgb}, V^{d/t} \rightarrow V^{rgb}, \alpha^{rgb} \rightarrow \alpha^{d/t}, \beta^{rgb} \rightarrow \beta^{d/t}$.

\begin{table*}[!t]
  \centering
  \caption{
    Comparison with recent state-of-the-art RGB-D SOD methods on NJUD~\cite{NJUD}, NLPR~\cite{NLPR}, SIP~\cite{SIP} and STEREO1000~\cite{STEREO}.
    $\star$: using the multi-scale training technique.
    \blank: not available.
    The best three results are highlighted using {\color{reda} \textbf{red}}, {\color{mygreen} \textbf{green}} and {\color{myblue} \textbf{blue}} in the order.
  }
  \label{tab:cmp_0}
  \setlength\tabcolsep{0.5em}
  \resizebox{0.95\linewidth}{!}{%
    \rowcolors{2}{gray!10}{white}
    \begin{tabular}{l|*5{c}|*5{c}|*5{c}|*5{c}}
  \toprule[2pt]
                                                 & \multicolumn{5}{c}{\textbf{NJUD}} & \multicolumn{5}{c}{\textbf{NLPR}} & \multicolumn{5}{c}{\textbf{SIP}} & \multicolumn{5}{c}{\textbf{STEREO1000}}                                                                                                                                                                                                                                                                                                                                                                                                                                                                                                                                                                                 \\
  \multirow{-2}{*}{\textbf{METHOD}}              & $S_{m}~\uparrow$                  & $F^{\omega}_{\beta}~\uparrow$     & $MAE~\downarrow$                 & $F_{\beta}~\uparrow$                    & $E_{m}~\uparrow$                 & $S_{m}~\uparrow$                 & $F^{\omega}_{\beta}~\uparrow$    & $MAE~\downarrow$                 & $F_{\beta}~\uparrow$             & $E_{m}~\uparrow$                 & $S_{m}~\uparrow$                 & $F^{\omega}_{\beta}~\uparrow$    & $MAE~\downarrow$                 & $F_{\beta}~\uparrow$             & $E_{m}~\uparrow$                 & $S_{m}~\uparrow$                 & $F^{\omega}_{\beta}~\uparrow$    & $MAE~\downarrow$                 & $F_{\beta}~\uparrow$             & $E_{m}~\uparrow$                 \\
  \midrule[1pt]
  CPFP$_{19}$~\cite{CPFP}                        & 0.878                             & 0.828                             & 0.053                            & 0.877                                   & 0.900                            & 0.884                            & 0.807                            & 0.038                            & 0.862                            & 0.920                            & 0.850                            & 0.788                            & 0.064                            & 0.851                            & 0.899                            & 0.879                            & 0.817                            & 0.051                            & 0.874                            & 0.907                            \\
  DMRA$_{19}$~\cite{DUTRGBD}                     & 0.886                             & 0.846                             & 0.051                            & 0.886                                   & 0.920                            & 0.899                            & 0.838                            & 0.031                            & 0.879                            & 0.941                            & 0.806                            & 0.739                            & 0.086                            & 0.821                            & 0.863                            & 0.752                            & 0.647                            & 0.087                            & 0.743                            & 0.816                            \\
  MMCI$_{19}$~\cite{MMCI}                        & 0.859                             & 0.739                             & 0.079                            & 0.853                                   & 0.882                            & 0.856                            & 0.676                            & 0.059                            & 0.815                            & 0.872                            & 0.833                            & 0.712                            & 0.086                            & 0.818                            & 0.886                            & 0.873                            & 0.760                            & 0.068                            & 0.863                            & 0.905                            \\
  TANet$_{19}$~\cite{TANet}                      & 0.878                             & 0.803                             & 0.061                            & 0.874                                   & 0.909                            & 0.886                            & 0.779                            & 0.041                            & 0.863                            & 0.916                            & 0.835                            & 0.748                            & 0.075                            & 0.830                            & 0.894                            & 0.871                            & 0.787                            & 0.060                            & 0.861                            & 0.916                            \\
  JLDCF$_{20}$~\cite{JLDCF-RGBDSOD-journal}      & 0.902                             & 0.869                             & 0.041                            & 0.904                                   & 0.935                            & 0.925                            & 0.882                            & {\color{myblue} \textbf{0.022}}  & 0.918                            & 0.955                            & 0.880                            & 0.844                            & 0.049                            & 0.889                            & 0.923                            & 0.903                            & 0.857                            & 0.040                            & 0.904                            & 0.937                            \\
  S2MA$_{20}$~\cite{S2MA}                        & 0.894                             & 0.842                             & 0.053                            & 0.889                                   & 0.916                            & 0.915                            & 0.852                            & 0.030                            & 0.902                            & 0.942                            & 0.872                            & 0.819                            & 0.057                            & 0.877                            & 0.913                            & 0.890                            & 0.825                            & 0.051                            & 0.882                            & 0.926                            \\
  UCNet$_{20}$~\cite{UCNet-RGBDSOD-conference}   & 0.897                             & 0.868                             & 0.043                            & 0.895                                   & 0.934                            & 0.920                            & 0.878                            & 0.025                            & 0.903                            & 0.955                            & 0.875                            & 0.836                            & 0.051                            & 0.879                            & 0.918                            & 0.903                            & 0.867                            & 0.039                            & 0.899                            & 0.942                            \\
  BBSNet$_{20}$~\cite{BBSNet-RGBDSOD-journal}    & {\color{myblue} \textbf{0.926}}   & 0.892                             & 0.033                            & {\color{myblue} \textbf{0.929}}         & 0.944                            & {\color{myblue} \textbf{0.930}}  & 0.878                            & 0.024                            & 0.915                            & 0.952                            & 0.887                            & 0.840                            & 0.052                            & 0.895                            & 0.922                            & 0.913                            & 0.864                            & 0.039                            & 0.910                            & 0.941                            \\
  CMWNet$_{20}$~\cite{CMWNet}                    & 0.903                             & 0.857                             & 0.046                            & 0.902                                   & 0.923                            & 0.917                            & 0.856                            & 0.029                            & 0.903                            & 0.940                            & 0.867                            & 0.811                            & 0.062                            & 0.874                            & 0.909                            & 0.905                            & 0.847                            & 0.043                            & 0.901                            & 0.930                            \\
  CoNet$_{20}$~\cite{CoNet}                      & 0.896                             & 0.848                             & 0.046                            & 0.893                                   & 0.924                            & 0.908                            & 0.841                            & 0.031                            & 0.887                            & 0.933                            & 0.858                            & 0.802                            & 0.063                            & 0.867                            & 0.909                            & 0.905                            & 0.865                            & 0.038                            & 0.901                            & 0.941                            \\
  DANet$_{20}$~\cite{DANet}                      & 0.899                             & 0.857                             & 0.045                            & 0.898                                   & 0.922                            & 0.915                            & 0.862                            & 0.028                            & 0.903                            & 0.949                            & 0.875                            & 0.822                            & 0.054                            & 0.876                            & 0.915                            & 0.901                            & 0.846                            & 0.043                            & 0.892                            & 0.931                            \\
  HDFNet$_{20}$~\cite{HDFNet}                    & 0.908                             & 0.877                             & 0.038                            & 0.911                                   & 0.932                            & 0.923                            & 0.882                            & 0.023                            & 0.917                            & 0.957                            & 0.886                            & 0.848                            & 0.047                            & 0.894                            & 0.924                            & 0.900                            & 0.853                            & 0.041                            & 0.900                            & 0.931                            \\
  ICNet$_{20}$~\cite{ICNet}                      & 0.894                             & 0.843                             & 0.052                            & 0.891                                   & 0.913                            & 0.923                            & 0.864                            & 0.028                            & 0.908                            & 0.945                            & 0.854                            & 0.791                            & 0.069                            & 0.857                            & 0.900                            & 0.903                            & 0.844                            & 0.045                            & 0.898                            & 0.926                            \\
  $\star$D3Net$_{20}$~\cite{SIP}                 & 0.900                             & 0.854                             & 0.046                            & 0.900                                   & 0.916                            & 0.912                            & 0.849                            & 0.030                            & 0.897                            & 0.945                            & 0.860                            & 0.799                            & 0.063                            & 0.861                            & 0.902                            & 0.899                            & 0.838                            & 0.046                            & 0.891                            & 0.924                            \\
  SPNet$_{21}$~\cite{SPNet-RGBDSOD-journal}      & 0.924                             & {\color{myblue} \textbf{0.906}}   & {\color{reda} \textbf{0.028}}    & 0.928                                   & {\color{myblue} \textbf{0.953}}  & 0.927                            & 0.896                            & {\color{mygreen} \textbf{0.021}} & 0.919                            & 0.959                            & 0.894                            & 0.868                            & 0.043                            & 0.904                            & 0.931                            & 0.907                            & 0.873                            & 0.037                            & 0.906                            & 0.942                            \\
  RD3D$_{21}$~\cite{RD3D-RGBDSOD}                & 0.916                             & 0.886                             & 0.036                            & 0.914                                   & 0.942                            & {\color{myblue} \textbf{0.930}}  & 0.889                            & {\color{myblue} \textbf{0.022}}  & 0.919                            & 0.959                            & 0.885                            & 0.845                            & 0.048                            & 0.889                            & 0.924                            & 0.911                            & 0.871                            & 0.037                            & 0.906                            & 0.944                            \\
  TriTransNet$_{21}$~\cite{TriTransNet-RGBDSOD}  & 0.920                             & {\color{myblue} \textbf{0.906}}   & {\color{myblue} \textbf{0.030}}  & 0.926                                   & {\color{mygreen} \textbf{0.954}} & 0.928                            & {\color{mygreen} \textbf{0.902}} & {\color{reda} \textbf{0.020}}    & {\color{mygreen} \textbf{0.924}} & {\color{mygreen} \textbf{0.964}} & 0.886                            & 0.864                            & 0.043                            & 0.899                            & 0.929                            & 0.908                            & 0.882                            & {\color{mygreen} \textbf{0.033}} & 0.911                            & {\color{mygreen} \textbf{0.950}} \\
  DCF$_{21}$~\cite{DCF-RGBDSOD}                  & 0.904                             & 0.876                             & 0.039                            & 0.905                                   & 0.940                            & 0.922                            & 0.884                            & 0.024                            & 0.910                            & 0.956                            & 0.874                            & 0.840                            & 0.052                            & 0.886                            & 0.921                            & 0.906                            & 0.872                            & 0.037                            & 0.904                            & 0.943                            \\
  HAINet$_{21}$~\cite{HAINet-RGBDSOD}            & 0.912                             & 0.883                             & 0.038                            & 0.915                                   & 0.934                            & 0.924                            & 0.887                            & 0.024                            & 0.915                            & 0.959                            & 0.880                            & 0.842                            & 0.053                            & 0.892                            & 0.919                            & 0.907                            & 0.866                            & 0.040                            & 0.906                            & 0.935                            \\
  CCAFNet$_{21}$~\cite{CCAFNet-RGBDSOD}          & 0.910                             & 0.877                             & 0.037                            & 0.910                                   & 0.941                            & 0.922                            & 0.875                            & 0.027                            & 0.909                            & 0.952                            & 0.877                            & 0.829                            & 0.054                            & 0.880                            & 0.916                            & 0.892                            & 0.844                            & 0.045                            & 0.887                            & 0.932                            \\
  DCMF$_{21}$~\cite{DCMF-RGBDSOD}                & 0.913                             & 0.867                             & 0.043                            & 0.915                                   & 0.925                            & 0.922                            & 0.856                            & 0.029                            & 0.906                            & 0.940                            & 0.870                            & 0.808                            & 0.062                            & 0.872                            & 0.906                            & 0.910                            & 0.849                            & 0.043                            & 0.906                            & 0.930                            \\
  UCNet-CVAE$_{21}$~\cite{UCNet-RGBDSOD-journal} & 0.904                             & 0.886                             & 0.038                            & 0.907                                   & 0.943                            & 0.922                            & 0.889                            & 0.023                            & 0.909                            & 0.956                            & 0.882                            & 0.850                            & 0.045                            & 0.889                            & 0.927                            & 0.906                            & 0.878                            & {\color{myblue} \textbf{0.036}}  & 0.904                            & 0.945                            \\
  $\star$CMINet$_{21}$~\cite{COME15K-CMINet}     & {\color{reda} \textbf{0.929}}     & {\color{reda} \textbf{0.910}}     & {\color{mygreen} \textbf{0.029}} & {\color{reda} \textbf{0.934}}           & {\color{myblue} \textbf{0.953}}  & {\color{mygreen} \textbf{0.932}} & {\color{myblue} \textbf{0.900}}  & {\color{mygreen} \textbf{0.021}} & 0.922                            & {\color{myblue} \textbf{0.962}}  & {\color{mygreen} \textbf{0.899}} & {\color{mygreen} \textbf{0.872}} & {\color{mygreen} \textbf{0.040}} & {\color{mygreen} \textbf{0.910}} & {\color{mygreen} \textbf{0.937}} & {\color{reda} \textbf{0.918}}    & {\color{mygreen} \textbf{0.886}} & {\color{reda} \textbf{0.032}}    & {\color{mygreen} \textbf{0.916}} & 0.948                            \\
  \midrule[1pt]
  OursR50(I)                                     & 0.921                             & 0.901                             & 0.031                            & 0.925                                   & {\color{myblue} \textbf{0.953}}  & 0.929                            & 0.895                            & {\color{reda} \textbf{0.020}}    & 0.921                            & {\color{myblue} \textbf{0.962}}  & 0.893                            & 0.864                            & {\color{myblue} \textbf{0.042}}  & 0.902                            & 0.933                            & 0.913                            & 0.882                            & {\color{mygreen} \textbf{0.033}} & {\color{myblue} \textbf{0.912}}  & {\color{myblue} \textbf{0.949}}  \\
  OursR50(II)                                    & 0.920                             & 0.900                             & 0.031                            & 0.923                                   & 0.951                            & 0.929                            & 0.895                            & {\color{myblue} \textbf{0.022}}  & 0.921                            & 0.961                            & 0.893                            & 0.868                            & {\color{myblue} \textbf{0.042}}  & 0.906                            & 0.933                            & {\color{myblue} \textbf{0.914}}  & {\color{myblue} \textbf{0.883}}  & {\color{mygreen} \textbf{0.033}} & 0.911                            & {\color{myblue} \textbf{0.949}}  \\
  OursR101(I)                                    & {\color{mygreen} \textbf{0.927}}  & {\color{mygreen} \textbf{0.909}}  & {\color{reda} \textbf{0.028}}    & {\color{mygreen} \textbf{0.932}}        & {\color{reda} \textbf{0.956}}    & {\color{myblue} \textbf{0.930}}  & 0.898                            & 0.023                            & {\color{myblue} \textbf{0.923}}  & 0.961                            & {\color{myblue} \textbf{0.895}}  & {\color{myblue} \textbf{0.871}}  & 0.043                            & {\color{myblue} \textbf{0.909}}  & {\color{myblue} \textbf{0.935}}  & {\color{mygreen} \textbf{0.917}} & {\color{reda} \textbf{0.888}}    & {\color{reda} \textbf{0.032}}    & {\color{reda} \textbf{0.918}}    & {\color{reda} \textbf{0.951}}    \\
  OursR101(II)                                   & {\color{myblue} \textbf{0.926}}   & {\color{myblue} \textbf{0.906}}   & {\color{mygreen} \textbf{0.029}} & 0.928                                   & {\color{myblue} \textbf{0.953}}  & {\color{reda} \textbf{0.934}}    & {\color{reda} \textbf{0.904}}    & {\color{mygreen} \textbf{0.021}} & {\color{reda} \textbf{0.929}}    & {\color{reda} \textbf{0.966}}    & {\color{reda} \textbf{0.904}}    & {\color{reda} \textbf{0.879}}    & {\color{reda} \textbf{0.037}}    & {\color{reda} \textbf{0.915}}    & {\color{reda} \textbf{0.943}}    & {\color{mygreen} \textbf{0.917}} & {\color{reda} \textbf{0.888}}    & {\color{reda} \textbf{0.032}}    & {\color{mygreen} \textbf{0.916}} & {\color{reda} \textbf{0.951}}    \\
  \bottomrule[2pt]
\end{tabular}

  }
\end{table*}

\begin{table*}[!t]
  \centering
  \caption{
    Comparison with recent state-of-the-art RGB-D SOD methods on SSD~\cite{SSD}, LFSD~\cite{LFSD} and DUTRGBD~\cite{DUTRGBD}.
    $\star$: using the multi-scale training technique.
    \blank: not available.
    The best three results are highlighted using {\color{reda} \textbf{red}}, {\color{mygreen} \textbf{green}} and {\color{myblue} \textbf{blue}} in the order.
  }
  \label{tab:cmp_1}
  \setlength\tabcolsep{0.5em}
  \resizebox{0.65\linewidth}{!}{%
    \rowcolors{2}{gray!10}{white}
    \begin{tabular}{l|*5{c}|*5{c}|*5{c}}
  \toprule[2pt]
                                                 & \multicolumn{5}{c}{\textbf{SSD}} & \multicolumn{5}{c}{\textbf{LFSD}} & \multicolumn{5}{c}{\textbf{DUTRGBD}}                                                                                                                                                                                                                                                                                                                                                                                                                                     \\
  \multirow{-2}{*}{\textbf{METHOD}}              & $S_{m}~\uparrow$                 & $F^{\omega}_{\beta}~\uparrow$     & $MAE~\downarrow$                     & $F_{\beta}~\uparrow$             & $E_{m}~\uparrow$                 & $S_{m}~\uparrow$                 & $F^{\omega}_{\beta}~\uparrow$    & $MAE~\downarrow$                 & $F_{\beta}~\uparrow$             & $E_{m}~\uparrow$                 & $S_{m}~\uparrow$                 & $F^{\omega}_{\beta}~\uparrow$    & $MAE~\downarrow$                 & $F_{\beta}~\uparrow$             & $E_{m}~\uparrow$                 \\
  \midrule[1pt]
  CPFP$_{19}$~\cite{CPFP}                        & 0.807                            & 0.708                             & 0.082                                & 0.766                            & 0.832                            & 0.828                            & 0.775                            & 0.088                            & 0.826                            & 0.867                            & 0.749                            & 0.637                            & 0.100                            & 0.718                            & 0.815                            \\
  DMRA$_{19}$~\cite{DUTRGBD}                     & 0.857                            & 0.784                             & 0.059                                & 0.844                            & 0.898                            & 0.847                            & 0.811                            & 0.076                            & 0.856                            & 0.899                            & 0.888                            & 0.851                            & 0.048                            & 0.897                            & 0.930                            \\
  MMCI$_{19}$~\cite{MMCI}                        & 0.814                            & 0.661                             & 0.082                                & 0.782                            & 0.860                            & 0.787                            & 0.663                            & 0.132                            & 0.771                            & 0.840                            & 0.791                            & 0.627                            & 0.112                            & 0.767                            & 0.856                            \\
  TANet$_{19}$~\cite{TANet}                      & 0.839                            & 0.726                             & 0.063                                & 0.810                            & 0.886                            & 0.801                            & 0.719                            & 0.111                            & 0.796                            & 0.851                            & 0.808                            & 0.704                            & 0.093                            & 0.790                            & 0.871                            \\
  JLDCF$_{20}$~\cite{JLDCF-RGBDSOD-journal}      & 0.860                            & 0.782                             & 0.053                                & 0.833                            & 0.899                            & 0.861                            & 0.822                            & 0.070                            & 0.867                            & 0.902                            & 0.905                            & 0.863                            & 0.043                            & 0.911                            & 0.938                            \\
  S2MA$_{20}$~\cite{S2MA}                        & 0.868                            & 0.787                             & 0.052                                & 0.848                            & 0.898                            & 0.837                            & 0.772                            & 0.094                            & 0.835                            & 0.876                            & 0.903                            & 0.862                            & 0.044                            & 0.900                            & 0.935                            \\
  UCNet$_{20}$~\cite{UCNet-RGBDSOD-conference}   & 0.866                            & 0.813                             & 0.049                                & 0.854                            & 0.901                            & 0.864                            & 0.832                            & 0.066                            & 0.864                            & 0.906                            & 0.864                            & 0.820                            & 0.057                            & 0.857                            & 0.906                            \\
  BBSNet$_{20}$~\cite{BBSNet-RGBDSOD-journal}    & 0.868                            & 0.789                             & 0.051                                & 0.842                            & 0.904                            & {\color{myblue} \textbf{0.878}}  & 0.826                            & 0.065                            & 0.873                            & 0.907                            & 0.920                            & 0.883                            & 0.037                            & 0.927                            & 0.949                            \\
  CMWNet$_{20}$~\cite{CMWNet}                    & 0.875                            & 0.795                             & 0.051                                & 0.871                            & 0.902                            & 0.876                            & 0.834                            & 0.066                            & {\color{mygreen} \textbf{0.883}} & 0.908                            & 0.887                            & 0.831                            & 0.056                            & 0.888                            & 0.922                            \\
  CoNet$_{20}$~\cite{CoNet}                      & 0.853                            & 0.779                             & 0.060                                & 0.840                            & 0.898                            & 0.862                            & 0.814                            & 0.071                            & 0.859                            & 0.901                            & 0.919                            & 0.890                            & 0.034                            & 0.927                            & 0.952                            \\
  DANet$_{20}$~\cite{DANet}                      & 0.864                            & 0.795                             & 0.050                                & 0.843                            & 0.911                            & 0.849                            & 0.795                            & 0.079                            & 0.844                            & 0.881                            & 0.899                            & 0.860                            & 0.043                            & 0.906                            & 0.937                            \\
  HDFNet$_{20}$~\cite{HDFNet}                    & 0.879                            & 0.821                             & 0.045                                & 0.870                            & 0.911                            & 0.854                            & 0.806                            & 0.076                            & 0.862                            & 0.891                            & 0.907                            & 0.864                            & 0.041                            & 0.918                            & 0.938                            \\
  ICNet$_{20}$~\cite{ICNet}                      & 0.848                            & 0.772                             & 0.064                                & 0.841                            & 0.879                            & 0.868                            & 0.822                            & 0.071                            & 0.871                            & 0.900                            & 0.852                            & 0.784                            & 0.072                            & 0.850                            & 0.901                            \\
  $\star$D3Net$_{20}$~\cite{SIP}                 & 0.857                            & 0.777                             & 0.058                                & 0.834                            & 0.904                            & 0.825                            & 0.760                            & 0.095                            & 0.810                            & 0.863                            & 0.775                            & 0.668                            & 0.097                            & 0.742                            & 0.849                            \\
  SPNet$_{21}$~\cite{SPNet-RGBDSOD-journal}      & 0.871                            & 0.823                             & 0.044                                & 0.863                            & 0.920                            & 0.854                            & 0.823                            & 0.071                            & 0.863                            & 0.897                            & 0.804                            & 0.735                            & 0.085                            & 0.849                            & 0.877                            \\
  RD3D$_{21}$~\cite{RD3D-RGBDSOD}                & 0.803                            & 0.707                             & 0.082                                & 0.772                            & 0.869                            & 0.858                            & 0.816                            & 0.073                            & 0.854                            & 0.898                            & {\color{myblue} \textbf{0.931}}  & 0.907                            & 0.031                            & {\color{myblue} \textbf{0.939}}  & 0.957                            \\
  TriTransNet$_{21}$~\cite{TriTransNet-RGBDSOD}  & 0.881                            & {\color{mygreen} \textbf{0.842}}  & {\color{myblue} \textbf{0.041}}      & {\color{myblue} \textbf{0.873}}  & {\color{mygreen} \textbf{0.935}} & 0.866                            & 0.840                            & 0.066                            & 0.870                            & 0.908                            & {\color{mygreen} \textbf{0.933}} & {\color{reda} \textbf{0.926}}    & {\color{reda} \textbf{0.025}}    & {\color{reda} \textbf{0.946}}    & {\color{mygreen} \textbf{0.966}} \\
  DCF$_{21}$~\cite{DCF-RGBDSOD}                  & 0.852                            & 0.789                             & 0.054                                & 0.829                            & 0.905                            & 0.856                            & 0.823                            & 0.071                            & 0.860                            & 0.903                            & 0.924                            & {\color{myblue} \textbf{0.909}}  & 0.030                            & 0.932                            & 0.957                            \\
  HAINet$_{21}$~\cite{HAINet-RGBDSOD}            & 0.857                            & 0.798                             & 0.052                                & 0.838                            & 0.908                            & 0.854                            & 0.811                            & 0.079                            & 0.853                            & 0.892                            & 0.910                            & 0.883                            & 0.038                            & 0.920                            & 0.939                            \\
  CCAFNet$_{21}$~\cite{CCAFNet-RGBDSOD}          & 0.863                            & 0.799                             & 0.048                                & 0.842                            & 0.916                            & 0.827                            & 0.783                            & 0.087                            & 0.832                            & 0.877                            & 0.904                            & 0.878                            & 0.038                            & 0.913                            & 0.943                            \\
  DCMF$_{21}$~\cite{DCMF-RGBDSOD}                & {\color{myblue} \textbf{0.882}}  & 0.803                             & 0.053                                & 0.867                            & 0.895                            & 0.877                            & 0.825                            & 0.068                            & 0.875                            & 0.905                            & 0.928                            & 0.888                            & 0.035                            & 0.932                            & 0.951                            \\
  UCNet-CVAE$_{21}$~\cite{UCNet-RGBDSOD-journal} & \blank                           & \blank                            & \blank                               & \blank                           & \blank                           & 0.864                            & 0.836                            & 0.064                            & 0.863                            & 0.908                            & \blank                           & \blank                           & \blank                           & \blank                           & \blank                           \\
  $\star$CMINet$_{21}$~\cite{COME15K-CMINet}     & 0.874                            & 0.820                             & 0.047                                & 0.860                            & 0.910                            & {\color{mygreen} \textbf{0.879}} & {\color{myblue} \textbf{0.845}}  & {\color{mygreen} \textbf{0.061}} & 0.874                            & {\color{myblue} \textbf{0.910}}  & 0.897                            & 0.867                            & 0.046                            & 0.891                            & 0.937                            \\
  \midrule[1pt]
  OursR50(I)                                     & 0.878                            & 0.824                             & {\color{myblue} \textbf{0.041}}      & 0.859                            & 0.920                            & 0.873                            & 0.842                            & {\color{myblue} \textbf{0.063}}  & 0.877                            & {\color{mygreen} \textbf{0.914}} & 0.903                            & 0.874                            & 0.042                            & 0.904                            & 0.937                            \\
  OursR50(II)                                    & 0.874                            & 0.819                             & 0.043                                & 0.854                            & 0.924                            & {\color{reda} \textbf{0.882}}    & {\color{reda} \textbf{0.854}}    & {\color{reda} \textbf{0.056}}    & {\color{reda} \textbf{0.886}}    & {\color{reda} \textbf{0.921}}    & {\color{myblue} \textbf{0.931}}  & {\color{mygreen} \textbf{0.918}} & {\color{myblue} \textbf{0.028}}  & {\color{mygreen} \textbf{0.942}} & {\color{myblue} \textbf{0.964}}  \\
  OursR101(I)                                    & {\color{mygreen} \textbf{0.887}} & {\color{myblue} \textbf{0.838}}   & {\color{reda} \textbf{0.037}}        & {\color{mygreen} \textbf{0.876}} & {\color{myblue} \textbf{0.932}}  & 0.863                            & 0.825                            & 0.074                            & 0.865                            & 0.907                            & 0.913                            & 0.886                            & 0.039                            & 0.920                            & 0.947                            \\
  OursR101(II)                                   & {\color{reda} \textbf{0.890}}    & {\color{reda} \textbf{0.849}}     & {\color{mygreen} \textbf{0.038}}     & {\color{reda} \textbf{0.884}}    & {\color{reda} \textbf{0.936}}    & 0.876                            & {\color{mygreen} \textbf{0.847}} & {\color{mygreen} \textbf{0.061}} & {\color{myblue} \textbf{0.880}}  & {\color{mygreen} \textbf{0.914}} & {\color{reda} \textbf{0.937}}    & {\color{reda} \textbf{0.926}}    & {\color{mygreen} \textbf{0.026}} & {\color{reda} \textbf{0.946}}    & {\color{reda} \textbf{0.967}}    \\
  \bottomrule[2pt]
\end{tabular}

  }
\end{table*}

\begin{table}[!t]
  \centering
  \caption{
    Comparison with recent state-of-the-art RGB-T SOD methods on VT821~\cite{VT821-MTMR}, VT1000~\cite{VT1000-SDGL} and VT5000-TE~\cite{VT5000-ADF}.
    The best three results are highlighted using {\color{reda} \textbf{red}}, {\color{mygreen} \textbf{green}} and {\color{myblue} \textbf{blue}} in the order.
  }
  \label{tab:cmp_2}
  \setlength\tabcolsep{0.1em}
  \resizebox{\linewidth}{!}{%
    \rowcolors{2}{gray!10}{white}
    \begin{tabular}{cc|*{10}{c}}
	\toprule[2pt]
	                                                                        &                              & MTMR$_{18}$            & M3S-NIR$_{19}$    & ADF$_{20}$         & SDGL$_{20}$         & MIDD$_{21}$           & CSRNet$_{21}$                   & CGFNet$_{21}$                   & ECFFNet$_{21}$                  & Ours$^{50}$                      & Ours$^{101}$                     \\
	\multicolumn{2}{c|}{\multirow{-2}{*}{\textbf{METHOD}}}                  & \cite{VT821-MTMR}            & \cite{M3S-NIR-RGBTSOD} & \cite{VT5000-ADF} & \cite{VT1000-SDGL} & \cite{MIDD-RGBTSOD} & \cite{CSRNet-RGBTSOD} & \cite{CGFNet-RGBTSOD}           & \cite{ECFFNet-RGBTSOD}          &                                 &                                                                     \\ \midrule[1pt]
	                                                                        & $S_{m}\uparrow$              & 0.593                  & 0.723             & 0.810              & 0.765               & 0.871                 & {\color{myblue} \textbf{0.885}} & 0.880                           & 0.877                           & {\color{mygreen} \textbf{0.891}} & {\color{reda} \textbf{0.898}}    \\
	                                                                        & $F^{\omega}_{\beta}\uparrow$ & 0.264                  & 0.407             & 0.626              & 0.583               & 0.760                 & 0.821                           & {\color{myblue} \textbf{0.829}} & 0.799                           & {\color{mygreen} \textbf{0.834}} & {\color{reda} \textbf{0.845}}    \\
	                                                                        & $M\downarrow$                & 0.260                  & 0.140             & 0.077              & 0.085               & 0.045                 & 0.038                           & 0.038                           & {\color{myblue} \textbf{0.035}} & {\color{mygreen} \textbf{0.033}} & {\color{reda} \textbf{0.027}}    \\
	                                                                        & $F_{\beta}\uparrow$          & 0.646                  & 0.738             & 0.752              & 0.735               & 0.851                 & 0.858                           & {\color{myblue} \textbf{0.866}} & 0.835                           & {\color{mygreen} \textbf{0.876}} & {\color{reda} \textbf{0.877}}    \\
	\multirow{-5}{*}{\rotatebox[]{90}{\textbf{VT821}}~\cite{VT821-MTMR}}    & $E_{m}\uparrow$              & 0.762                  & 0.861             & 0.845              & 0.847               & 0.898                 & 0.912                           & {\color{myblue} \textbf{0.918}} & 0.907                           & {\color{mygreen} \textbf{0.919}} & {\color{reda} \textbf{0.928}}    \\ \midrule[1pt]
	                                                                        & $S_{m}\uparrow$              & 0.706                  & 0.726             & 0.910              & 0.787               & 0.915                 & 0.918                           & 0.923                           & {\color{myblue} \textbf{0.924}} & {\color{mygreen} \textbf{0.936}} & {\color{reda} \textbf{0.938}}    \\
	                                                                        & $F^{\omega}_{\beta}\uparrow$ & 0.485                  & 0.463             & 0.804              & 0.652               & 0.856                 & 0.878                           & {\color{myblue} \textbf{0.900}} & 0.883                           & {\color{mygreen} \textbf{0.908}} & {\color{reda} \textbf{0.911}}    \\
	                                                                        & $M\downarrow$                & 0.119                  & 0.145             & 0.034              & 0.090               & 0.027                 & 0.024                           & 0.023                           & {\color{myblue} \textbf{0.022}} & {\color{mygreen} \textbf{0.018}} & {\color{reda} \textbf{0.017}}    \\
	                                                                        & $F_{\beta}\uparrow$          & 0.715                  & 0.735             & 0.908              & 0.770               & 0.913                 & 0.908                           & {\color{myblue} \textbf{0.923}} & 0.917                           & {\color{mygreen} \textbf{0.935}} & {\color{reda} \textbf{0.939}}    \\
	\multirow{-5}{*}{\rotatebox[]{90}{\textbf{VT1000}}~\cite{VT1000-SDGL}}  & $E_{m}\uparrow$              & 0.836                  & 0.828             & 0.922              & 0.857               & 0.942                 & 0.940                           & {\color{reda} \textbf{0.955}}   & {\color{myblue} \textbf{0.947}} & 0.945                            & {\color{mygreen} \textbf{0.949}} \\ \midrule[1pt]
	                                                                        & $S_{m}\uparrow$              & 0.680                  & 0.652             & 0.863              & 0.750               & 0.867                 & 0.868                           & {\color{myblue} \textbf{0.883}} & 0.875                           & {\color{mygreen} \textbf{0.892}} & {\color{reda} \textbf{0.899}}    \\
	                                                                        & $F^{\omega}_{\beta}\uparrow$ & 0.397                  & 0.327             & 0.722              & 0.558               & 0.763                 & 0.796                           & {\color{myblue} \textbf{0.831}} & 0.800                           & {\color{mygreen} \textbf{0.835}} & {\color{reda} \textbf{0.849}}    \\
	                                                                        & $M\downarrow$                & 0.114                  & 0.168             & 0.048              & 0.089               & 0.043                 & 0.042                           & {\color{myblue} \textbf{0.035}} & 0.038                           & {\color{mygreen} \textbf{0.032}} & {\color{reda} \textbf{0.028}}    \\
	                                                                        & $F_{\beta}\uparrow$          & 0.613                  & 0.596             & 0.837              & 0.695               & 0.849                 & 0.837                           & {\color{myblue} \textbf{0.869}} & 0.846                           & {\color{mygreen} \textbf{0.873}} & {\color{reda} \textbf{0.882}}    \\
	\multirow{-5}{*}{\rotatebox[]{90}{\textbf{VT5000TE}}~\cite{VT5000-ADF}} & $E_{m}\uparrow$              & 0.795                  & 0.782             & 0.891              & 0.824               & 0.899                 & 0.907                           & {\color{myblue} \textbf{0.924}} & 0.910                           & {\color{mygreen} \textbf{0.930}} & {\color{reda} \textbf{0.941}}    \\
	\bottomrule[2pt]
\end{tabular}

  }
\end{table}

\begin{figure*}[!t]
  \centering
  \subfloat[]{
    \includegraphics[width=0.54\linewidth]{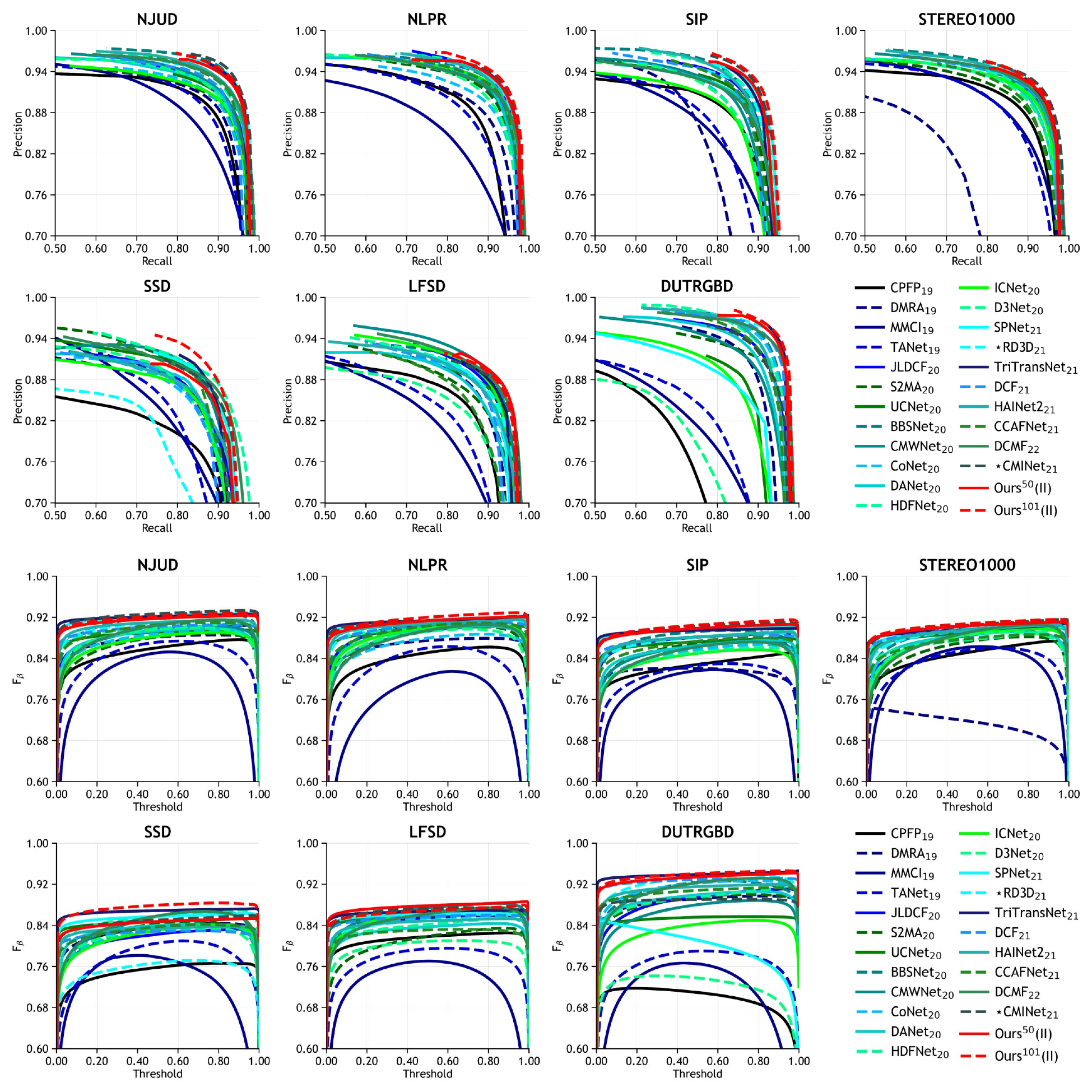}
    \label{fig:rgbd-prfm}
  }%
  \hfil
  \subfloat[]{
    \includegraphics[width=0.36\linewidth]{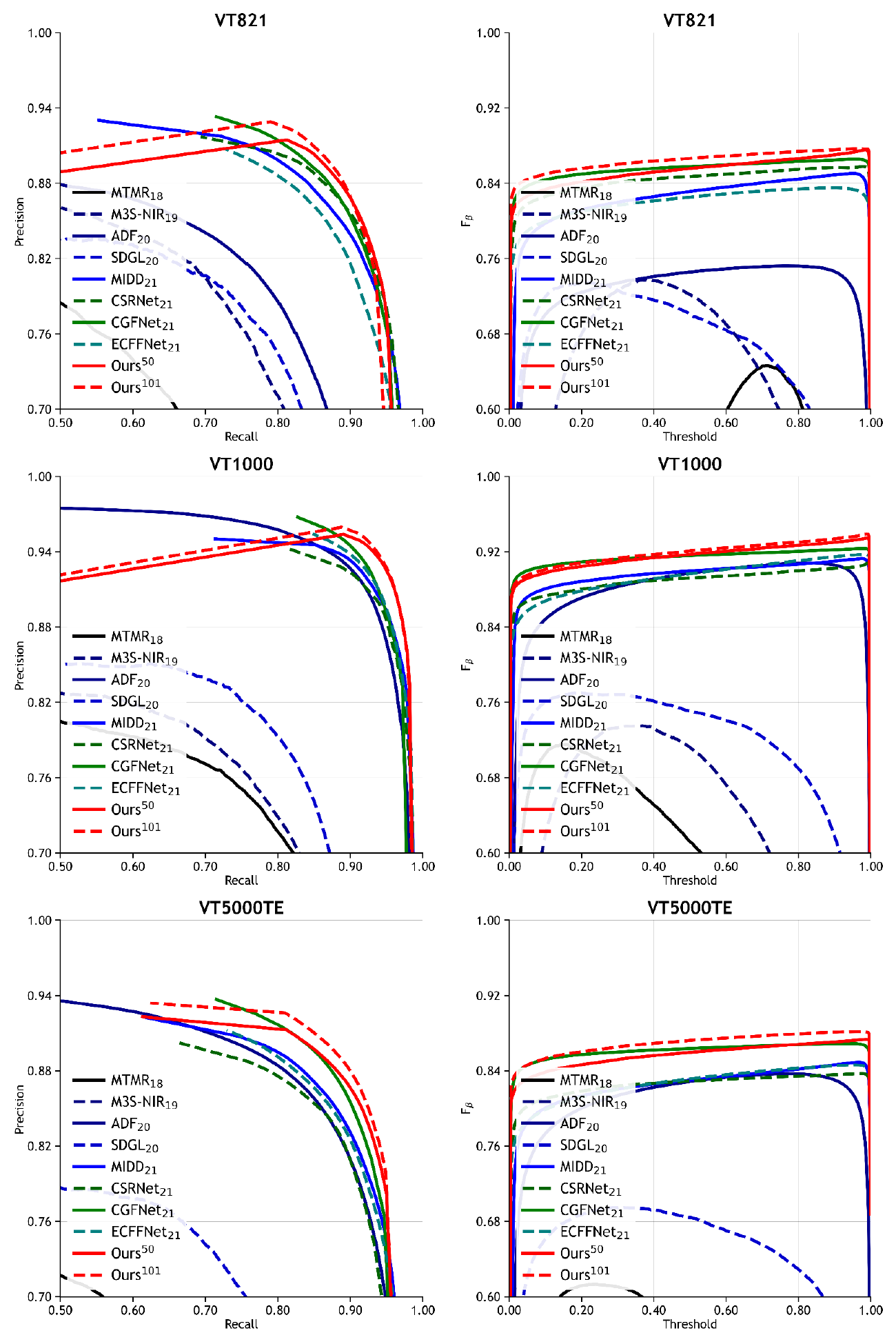}
    \label{fig:rgbt-prfm}
  }
  \caption{
    Precision-Recall (PR) and $F_{\beta}$-Threshold curves.
    $\star$: using the multi-scale training technique.
    (a) RGB-D SOD methods.
    (b) RGB-T SOD methods.
  }
  \label{fig:prfm}
\end{figure*}

\begin{table}[!t]
  \centering
  \caption{
    Comparison of FLOPs, the number of parameters, and FPS of some recent publicly available state-of-the-art RGB-D/RGB-T SOD methods.
    The evaluation process is performed on a single 2080Ti GPU while following the default inference settings of each method.
    $\ddagger$: Our self-attention block is replaced by the shifted window based self-attention~\cite{Swin}.
  }
  \label{tab:cmp_more}
  \resizebox{0.9\linewidth}{!}{%
    \rowcolors{2}{white}{gray!10}
    \begin{tabular}{r|r|c|c|c}
  \toprule[2pt]
  \textbf{METHOD}                               & \textbf{BACKBONE}                        & \textbf{FLOPs (G)} & \textbf{Params. (M)} & \textbf{FPS} \\ \midrule[1pt]
  SPNet$_{21}$~\cite{SPNet-RGBDSOD-journal}     & Res2Net-50-v1b-26w-4s~\cite{Res2Net}     & 135.857            & 175.291              & 26.871       \\
  RD3D$_{21}$~\cite{RD3D-RGBDSOD}               & I3DResNet-50~\cite{ResNet-I3D}           & 101.460            & 46.900               & 25.934       \\
  TriTransNet$_{21}$~\cite{TriTransNet-RGBDSOD} & ResNet-50~\cite{Resnet}+ViT-B~\cite{ViT} & 680.072            & 139.548              & 10.216       \\
  DCF$_{21}$~\cite{DCF-RGBDSOD}                 & ResNet-50~\cite{Resnet}                  & 107.815            & 108.491              & 23.869       \\
  HAINet$_{21}$~\cite{HAINet-RGBDSOD}           & VGG-16~\cite{VGG}                        & 350.831            & 59.823               & 12.187       \\
  CCAFNet$_{21}$~\cite{CCAFNet-RGBDSOD}         & VGG-16~\cite{VGG}                        & 153.064            & 41.798               & 64.154       \\
  DCMF$_{22}$~\cite{DCMF-RGBDSOD}               & VGG-16~\cite{VGG}                        & 271.058            & 58.937               & 23.093       \\
  CMINet$_{21}$~\cite{COME15K-CMINet}           & ResNet-50~\cite{Resnet}                  & 376.444            & 185.468              & 10.339       \\
  \midrule[0.5pt]
  MIDD$_{21}$~\cite{MIDD-RGBTSOD}               & VGG-16~\cite{VGG}                        & 434.437            & 52.428               & 21.526       \\
  CGFNet$_{21}$~\cite{CGFNet-RGBTSOD}           & VGG-16~\cite{VGG}                        & 760.556            & 66.382               & 12.538       \\
  \midrule[0.5pt]
  Ours$^{50}\ddagger$                           & ResNet-50d~\cite{ResNet-D}               & 44.392             & 55.515               & 26.623       \\
  Ours$^{50}$                                   & ResNet-50d~\cite{ResNet-D}               & 44.442             & 55.793               & 35.193       \\
  Ours$^{101}$                                  & ResNet-101d~\cite{ResNet-D}              & 63.907             & 93.777               & 27.666       \\
  \bottomrule[2pt]
\end{tabular}

  }
\end{table}

\section{Experiments}\label{sec:experiment}

\subsection{Datasets}\label{sec:dataset}

To validate the proposed model and components, we conducted experiments on seven RGB-D and three RGB-T benchmarks that have been widely used to evaluate RGB-D/RGB-T SOD methods.

\paragraph{RGB-D SOD.}
\textbf{NJUD}~\cite{NJUD} involves a lot of complex data, which consists of 1985 images collected from the Internet, 3D movies, and stereo photos, with their corresponding depth images.
\textbf{NLPR}~\cite{NLPR} contains 1000 pairs of RGB and depth images covering rich indoor and outdoor scenes.
\textbf{SIP}~\cite{SIP} containing 929 pairs of images is a recent high-resolution dataset. It is collected in an outdoor scene and contains complex lighting conditions and diverse human poses.
The 1000 stereoscopic images in \textbf{STEREO1000}~\cite{STEREO} were collected from Flickr, NVIDIA 3D Vision Live, and Stereoscopic Image Gallery.
\textbf{SSD}~\cite{SSD} is another RGB-D SOD dataset and only includes 80 samples covering indoor and outdoor scenes.
\textbf{LFSD}~\cite{LFSD} contains 100 pairs of RGB-D images and is built for saliency detection on the light field.
\textbf{DUTRGBD}~\cite{DUTRGBD} is a recently proposed large-scale RGB-D SOD dataset, which contains 1200 pairs of RGB-D images, 800 from indoors and 400 from outdoors. It captures a large number of challenging objects and scenes.
To make a fair comparison, we conduct experiments with two training settings.
(I): One is the setup in works~\cite{JLDCF-RGBDSOD-journal,S2MA,UCNet-RGBDSOD-journal,SIP,BBSNet-RGBDSOD-journal,CMWNet,SPNet-RGBDSOD-journal}, the training set only contains 1485 pairs from NJUD and 700 pairs from NLPR and the remaining data of these datasets as the test set.
(II): The other can be found in recent works~\cite{DCF-RGBDSOD,HAINet-RGBDSOD,TriTransNet-RGBDSOD,DUTRGBD,CoNet}. 700 images from NLPR, 1485 images from NJUD and 800 images from DUTRGBD are chosen as the training data.

\paragraph{RGB-T SOD.}
\textbf{VT821}~\cite{VT821-MTMR} includes 821 RGB-T image pairs and their ground truth annotations for the saliency detection purpose.
\textbf{VT1000}~\cite{VT1000-SDGL} contains 1000 pairs of RGB-T images including more than 400 kinds of common objects collected in 10 types of scenes under different illumination conditions.
\textbf{VT5000}~\cite{VT5000-ADF} is a large-scale dataset containing 5000 pairs of RGB-T images with ground truth annotations, which greatly improves the complexity and diversity of the scenes. It is split into VT5000TR (2500) and VT5000TE (2500).
Following the setting of recent works~\cite{MIDD-RGBTSOD,CSRNet-RGBTSOD,CGFNet-RGBTSOD,ECFFNet-RGBTSOD}, the training set only contains the 2500 samples from VT5000TR and all remaining data are used as the test set.

\subsection{Evaluation Metrics}~\label{sec:metric}

To fully demonstrate the performance differences between different methods, we introduced several metrics to quantitatively evaluate the models.
Specifically, \textbf{S-measure}~\cite{Smeasure} ($S_m$) focuses on region-aware and object-aware structural similarities between the saliency map and the ground truth.
\textbf{MAE}~\cite{MAE}, ($M$) indicates the average absolute pixel error.
\textbf{F-measure}~\cite{Fmeasure} ($F_{\beta}$) is a region-based similarity metric and based on precision and recall.
\textbf{E-measure}~\cite{Emeasure} ($E_{m}$) is characterized as both image-level statistics and local pixel matching.
\textbf{Weighted F-measure}~\cite{wFmeasure} ($F^{\omega}_{\beta}$) improves the metric $F_{\beta}$ by using a weighted precision for measuring exactness and a weighted recall for measuring completeness.
In addition to these metrics, we also introduce ``Precision-Recall'' and ``$F_{\beta}$-Threshold'' curves to present a comprehensive comparison of the model performance.

\subsection{Implementation Detail}

The backbone network is initialized by the weight pretrained on ImageNet, and the remaining structures are initialized randomly using the default method of the PyTorch toolbox.
All our models are trained for 100 epochs with a batch size of 8 using the SGD optimizer with a momentum of 0.9 and a weight decay of 0.0005 on an NVIDIA GTX 2080Ti GPU.
The learning rate is initialized as 0.005 and scheduled by the cosine strategy.
The single-channel depth/thermal input is repeated three times along the channel dimension to facilitate the use of pretrained parameters, and RGB and depth/thermal images are resized to $256 \times 256$.
Some data augmentation techniques are also introduced into the training phase to avoid over-fitting, such as some affine transforms, horizontal flipping and color jittering.
In the test stage, the RGB and depth/thermal images are resized to $256 \times 256$, and the final prediction is resized back to the original size for evaluation.
For all experiments, the model is supervised by the hybrid loss~\cite{BASNet}.

\begin{figure*}[!t]
  \centering
  \includegraphics[width=0.9\linewidth]{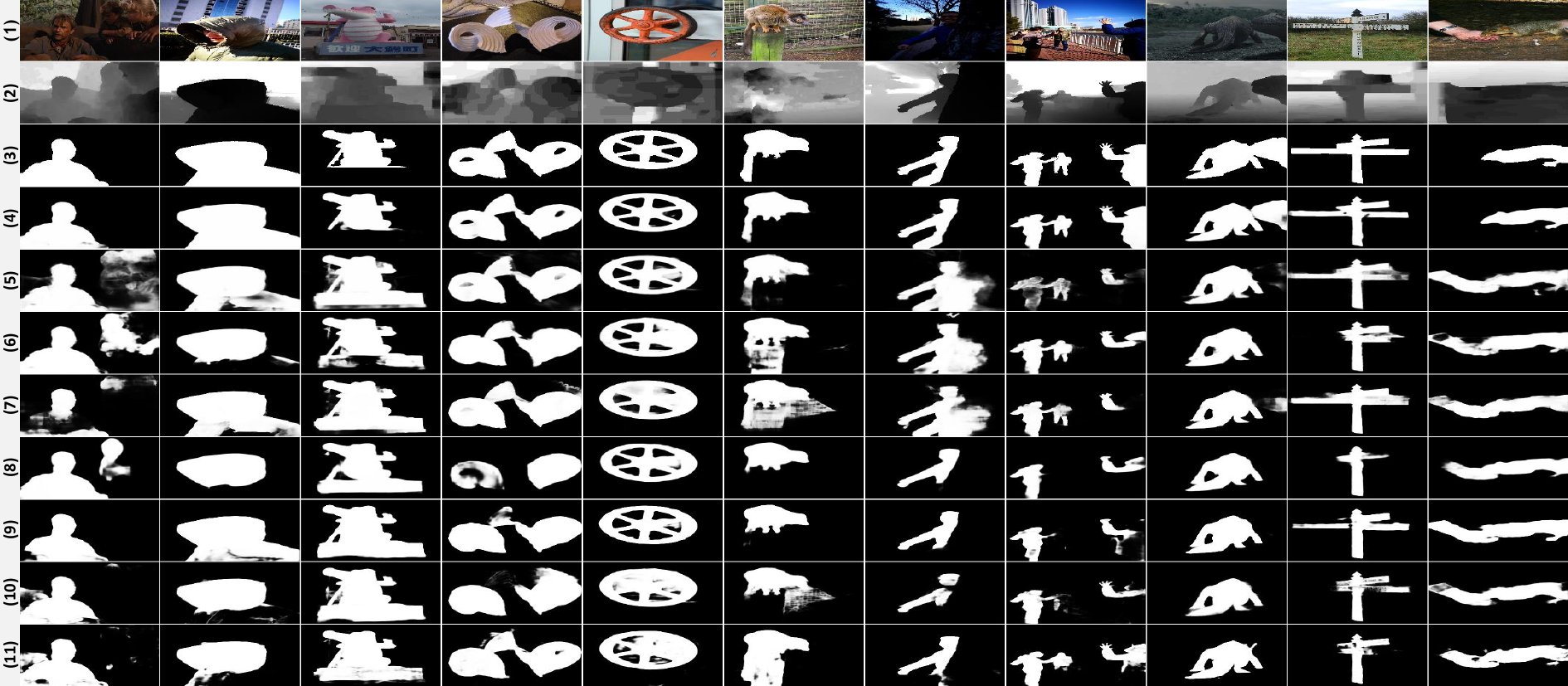}
  \caption{
    Qualitative comparison of our model with seven recent state-of-the-art models.
    (1): Image; (2) Depth; (3) Mask;
    (4): Ours$^{101}$(II);
    (5): DCMF~\cite{DCMF-RGBDSOD};
    (6): CCAFNet~\cite{CCAFNet-RGBDSOD};
    (7): HAINet~\cite{HAINet-RGBDSOD};
    (8): DCF~\cite{DCF-RGBDSOD};
    (9): TriTransNet~\cite{TriTransNet-RGBDSOD};
    (10): RD3D~\cite{RD3D-RGBDSOD};
    (11): SPNet~\cite{SPNet-RGBDSOD-journal}.
  }
  \label{fig:visualcmp}
\end{figure*}

\subsection{Comparison}\label{sec:comparison}

To demonstrate the effectiveness of our method, we compare it with the recent 31 state-of-the-art RGB-D and RGB-T SOD methods including CPFP~\cite{CPFP}, DMRA~\cite{DUTRGBD}, MMCI~\cite{MMCI}, TANet~\cite{TANet}, JLDCF~\cite{JLDCF-RGBDSOD-journal}, S2MA~\cite{S2MA}, UCNet~\cite{UCNet-RGBDSOD-conference}, BBSNet~\cite{BBSNet-RGBDSOD-journal}, CMWNet~\cite{CMWNet}, CoNet~\cite{CoNet}, DANet~\cite{DANet}, HDFNet~\cite{HDFNet}, ICNet~\cite{ICNet}, D3Net~\cite{SIP}, SPNet~\cite{SPNet-RGBDSOD-journal}, RD3D~\cite{RD3D-RGBDSOD}, TriTransNet~\cite{TriTransNet-RGBDSOD}, DCF~\cite{DCF-RGBDSOD}, HAINet~\cite{HAINet-RGBDSOD}, CCAFNet~\cite{CCAFNet-RGBDSOD}, UCNet-CVAE~\cite{UCNet-RGBDSOD-journal}, CMINet~\cite{COME15K-CMINet}, and DCMF~\cite{DCMF-RGBDSOD}, MTMR~\cite{VT821-MTMR}, M3S-NIR~\cite{M3S-NIR-RGBTSOD}, ADF~\cite{VT5000-ADF}, SDGL~\cite{VT1000-SDGL}, MIDD~\cite{MIDD-RGBTSOD}, CSRNet~\cite{CSRNet-RGBTSOD}, CGFNet~\cite{CGFNet-RGBTSOD} and ECFFNet~\cite{ECFFNet-RGBTSOD}.
All data used in experiments are from the resources released by the authors.

\paragraph{Quantitative Comparison.}\label{par:quantitative}
In Tab.~\ref{tab:cmp_0}, Tab.~\ref{tab:cmp_1} and Tab.~\ref{tab:cmp_2}, the detailed results from all competitors on ten datasets in two tasks and five metrics are listed and our methods perform best on all these datasets.
For RGB-D SOD, ``Ours$^{101}$(II)'' achieves the best average performance of 0.912 $S_{m}$, 0.886 $F^{\omega}_{\beta}$, 0.035 $M$, 0.947 $E_{m}$ and 0.914 $F_{\beta}$,
which are significant relative gains of 0.98\% $S_{m}$, 0.60\%$F^{\omega}_{\beta}$, 5.43\% $M$, 0.36\% $E_{m}$ and 	0.77\% $F_{\beta}$ over the second-best method TriTransNet~\cite{TriTransNet-RGBDSOD} which has more parameters, FLOPs and latency (see Tab.~\ref{tab:cmp_more}).
In the comparison on RGB-T SOD, our method also has more consistent and obvious performance gains.
It should be emphasized that this work focuses on exploring and designing a new architecture suitable for bi-modal SOD.
Therefore, the whole network is very simple and we do not explore collaboration with other modules, such as ASPP~\cite{Deeplab}, DenseASPP~\cite{DenseASPP}, and convolutional channel and spatial attention~\cite{SENet,CBAM,SKNet}, which deserves more attention in future work.
To compare the overall performance of different methods, we also show PR and $F_{\beta}$ curves in Tab.~\ref{fig:prfm}.
It can be seen that our methods correspond to the curves positioned more upward, which indicates that the proposed model performs better.

\paragraph{Qualitative Comparison.}\label{par:qualitative}
Some visual comparisons of different models are listed in Fig.~\ref{fig:visualcmp}, which covers the representative methods published recently.
As we can see that these samples have various types of scenes and objects, including
the large object with large internal variability (Column 1-5),
the out-of-bounds object (Column 1, 2, 8, 10, and 11),
the medium-sized object with unclear and complex boundaries in the low-brightness scene (Column 6-9),
the sample with a strong background interference (Column 1, 3, 6, and 11)
These results fully demonstrate the robustness of the proposed algorithm against different data, which can be attributed to the powerful information modeling capability of the proposed view-mixed attention mechanism.

\begin{table*}[!t]
  \centering
  \caption{
    Ablation analysis of the different components.
    $\ddagger$: The model variant where our self-attention block is replaced by the shifted window based self-attention~\cite{Swin}.
    ``(Lin.)'': The model variant where the Conv-FFN is replaced by the linear FFN.
    The best results are highlighted using {\color{reda} \textbf{red}}.
  }
  \label{tab:ablation}
  \resizebox{\linewidth}{!}{%
    \rowcolors{2}{gray!10}{white}
    \begin{tabular}{l*{4}{|*{5}{c}}}
	\toprule[2pt]
	                                          & \multicolumn{5}{c|}{\textbf{NJUD}} & \multicolumn{5}{c|}{\textbf{SIP}} & \multicolumn{5}{c|}{\textbf{STEREO1000}} & \multicolumn{5}{c}{\textbf{AVERAGE}}                                                                                                                                                                                                                                                                                                                                                                                                         \\
	\multirow{-2}{*}{\textbf{METHOD}}         & $S_{m}\uparrow$                    & $F^{\omega}_{\beta}\uparrow$      & $M\downarrow$                            & $E_{m}\uparrow$                      & $F_{\beta}\uparrow$ & $S_{m}\uparrow$ & $F^{\omega}_{\beta}\uparrow$ & $M\downarrow$ & $E_{m}\uparrow$ & $F_{\beta}\uparrow$ & $S_{m}\uparrow$ & $F^{\omega}_{\beta}\uparrow$ & $M\downarrow$ & $E_{m}\uparrow$ & $F_{\beta}\uparrow$ & $S_{m}\uparrow$               & $F^{\omega}_{\beta}\uparrow$  & $M\downarrow$                 & $E_{m}\uparrow$               & $F_{\beta}\uparrow$           \\ \midrule[1pt]
	Baseline                                  & 0.917                              & 0.891                             & 0.034                                    & 0.939                                & 0.922               & 0.883           & 0.847                        & 0.048         & 0.924           & 0.891               & 0.903           & 0.862                        & 0.039         & 0.935           & 0.899               & 0.901                         & 0.867                         & 0.040                         & 0.933                         & 0.904                         \\
	+IMCA                                     & 0.920                              & 0.897                             & 0.032                                    & 0.950                                & 0.923               & 0.886           & 0.853                        & 0.047         & 0.927           & 0.895               & 0.908           & 0.873                        & 0.035         & 0.945           & 0.907               & 0.905                         & 0.874                         & 0.038                         & 0.941                         & 0.908                         \\
	+IMCA+IMSA                                & 0.917                              & 0.895                             & 0.032                                    & 0.949                                & 0.923               & 0.889           & 0.858                        & 0.045         & 0.930           & 0.898               & 0.911           & 0.879                        & 0.034         & 0.949           & 0.909               & 0.906                         & 0.877                         & 0.037                         & 0.943                         & 0.910                         \\
	+IMCA+CSSA                                & 0.922                              & 0.901                             & 0.031                                    & 0.952                                & 0.926               & 0.892           & 0.863                        & 0.043         & 0.932           & 0.901               & 0.911           & 0.878                        & 0.034         & 0.947           & 0.906               & 0.908                         & 0.881                         & 0.036                         & 0.944                         & 0.911                         \\
	+IMCA+IMSA+CSSA                           & 0.921                              & 0.901                             & 0.031                                    & 0.953                                & 0.925               & 0.893           & 0.864                        & 0.042         & 0.933           & 0.902               & 0.913           & 0.882                        & 0.033         & 0.949           & 0.912               & {\color{reda} \textbf{0.909}} & {\color{reda} \textbf{0.882}} & {\color{reda} \textbf{0.035}} & {\color{reda} \textbf{0.945}} & {\color{reda} \textbf{0.913}} \\
	+IMCA+IMSA$^{\ddagger}$+CSSA$^{\ddagger}$ & 0.921                              & 0.902                             & 0.030                                    & 0.954                                & 0.926               & 0.884           & 0.852                        & 0.047         & 0.925           & 0.891               & 0.909           & 0.876                        & 0.036         & 0.946           & 0.906               & 0.905                         & 0.877                         & 0.038                         & 0.942                         & 0.908                         \\
	+IMCA+IMSA+CSSA (Lin.)                    & 0.918                              & 0.897                             & 0.033                                    & 0.949                                & 0.923               & 0.885           & 0.851                        & 0.046         & 0.928           & 0.893               & 0.910           & 0.876                        & 0.035         & 0.948           & 0.908               & 0.904                         & 0.875                         & 0.038                         & 0.942                         & 0.908                         \\
	\bottomrule[2pt]
\end{tabular}

  }
\end{table*}

\begin{figure}[!t]
  \centering
  \includegraphics[width=\linewidth]{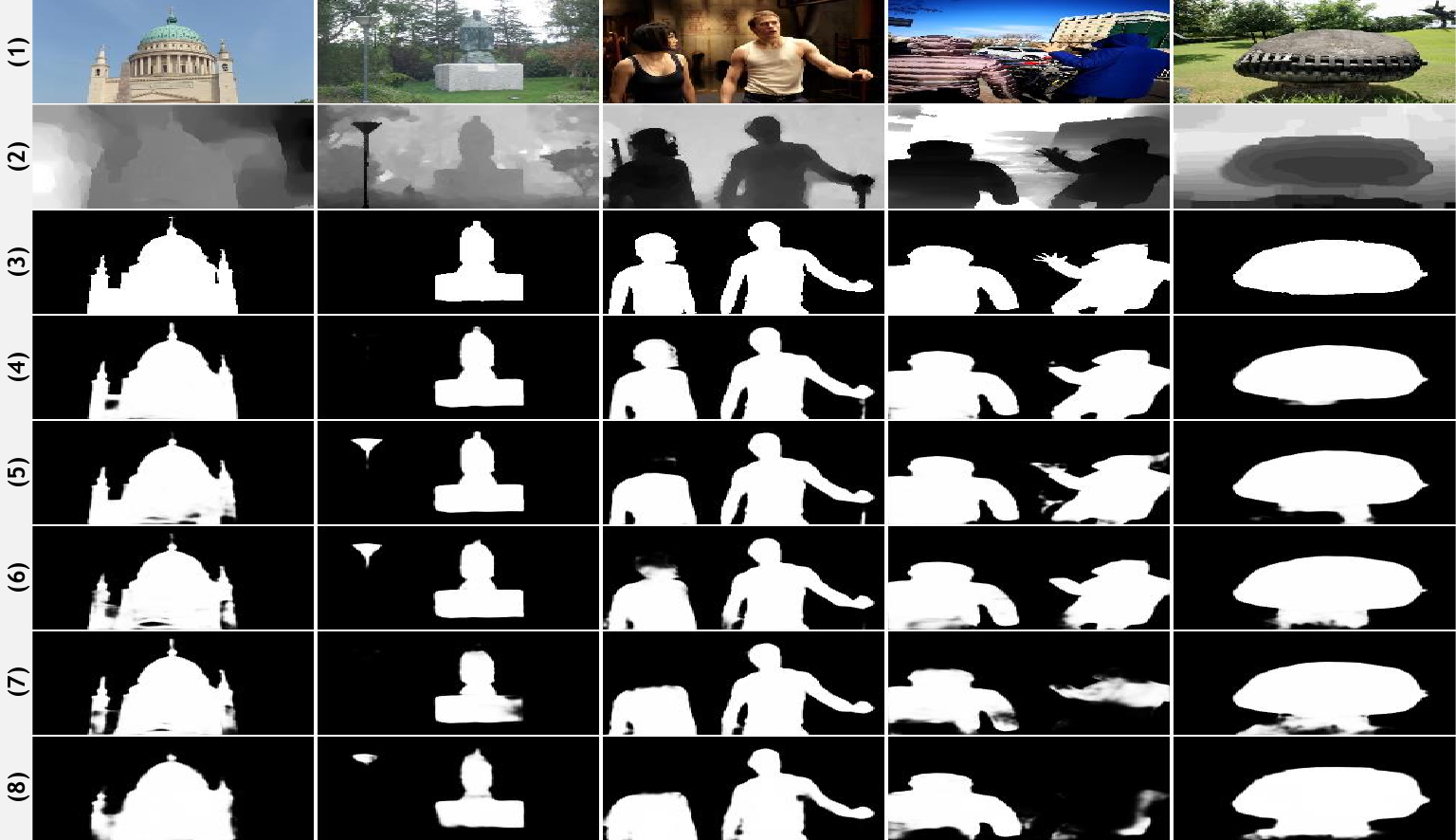}
  \caption{
    Visual ablation experiments for different components.
    (1): Image; (2) Depth; (3) Mask;
    (4): +IMCA+IMSA+CSSA;
    (5): +IMCA+CSSA;
    (6): +IMCA+IMSA;
    (7): +IMCA;
    (8): Baseline.
  }
  \label{fig:visualablation}
\end{figure}

\begin{table*}[!t]
  \centering
  \caption{
    Ablation analysis of the channel and spatial attention branches in the view-mixed attention block.
    $\times 0$ is equivalent to the branch being removed from the block.
  }
  \label{tab:casaattn_ffn}
  \resizebox{\linewidth}{!}{%
    \rowcolors{2}{gray!10}{white}
    \begin{tabular}{cc*{4}{|*{5}{c}}}
	\toprule[2pt]
	                                   &                                    & \multicolumn{5}{c|}{\textbf{NJUD}} & \multicolumn{5}{c|}{\textbf{SIP}} & \multicolumn{5}{c|}{\textbf{STEREO1000}} & \multicolumn{5}{c}{\textbf{AVERAGE}}                                                                                                                                                                                                                                                                                                                                                                                                         \\
	\multirow{-2}{*}{\textbf{SPATIAL}} & \multirow{-2}{*}{\textbf{CHANNEL}} & $S_{m}\uparrow$                    & $F^{\omega}_{\beta}\uparrow$      & $M\downarrow$                            & $E_{m}\uparrow$                      & $F_{\beta}\uparrow$ & $S_{m}\uparrow$ & $F^{\omega}_{\beta}\uparrow$ & $M\downarrow$ & $E_{m}\uparrow$ & $F_{\beta}\uparrow$ & $S_{m}\uparrow$ & $F^{\omega}_{\beta}\uparrow$ & $M\downarrow$ & $E_{m}\uparrow$ & $F_{\beta}\uparrow$ & $S_{m}\uparrow$               & $F^{\omega}_{\beta}\uparrow$  & $M\downarrow$                 & $E_{m}\uparrow$               & $F_{\beta}\uparrow$           \\ \midrule[1pt]
	$\times \alpha$                    & $\times \beta$                     & 0.921                              & 0.901                             & 0.031                                    & 0.953                                & 0.925               & 0.893           & 0.864                        & 0.042         & 0.933           & 0.902               & 0.913           & 0.882                        & 0.033         & 0.949           & 0.912               & {\color{reda} \textbf{0.909}} & {\color{reda} \textbf{0.882}} & {\color{reda} \textbf{0.035}} & {\color{reda} \textbf{0.945}} & {\color{reda} \textbf{0.913}} \\
	$\times 0.5$                       & $\times 0.5$                       & 0.920                              & 0.900                             & 0.030                                    & 0.952                                & 0.923               & 0.891           & 0.865                        & 0.044         & 0.932           & 0.904               & 0.912           & 0.880                        & 0.034         & 0.947           & 0.910               & 0.908                         & {\color{reda} \textbf{0.882}} & 0.036                         & 0.944                         & 0.912                         \\
	$\times 0$                         & $\times 1$                         & 0.918                              & 0.898                             & 0.033                                    & 0.949                                & 0.921               & 0.886           & 0.856                        & 0.046         & 0.928           & 0.897               & 0.910           & 0.877                        & 0.034         & 0.948           & 0.907               & 0.905                         & 0.877                         & 0.038                         & 0.942                         & 0.908                         \\
	$\times 1$                         & $\times 0$                         & 0.922                              & 0.902                             & 0.031                                    & 0.953                                & 0.926               & 0.889           & 0.858                        & 0.046         & 0.928           & 0.897               & 0.910           & 0.878                        & 0.034         & 0.947           & 0.909               & 0.907                         & 0.879                         & 0.037                         & 0.943                         & 0.911                         \\
	\bottomrule[2pt]
\end{tabular}

  }
\end{table*}
\begin{figure*}[!t]
  \centering
  \includegraphics[width=\linewidth]{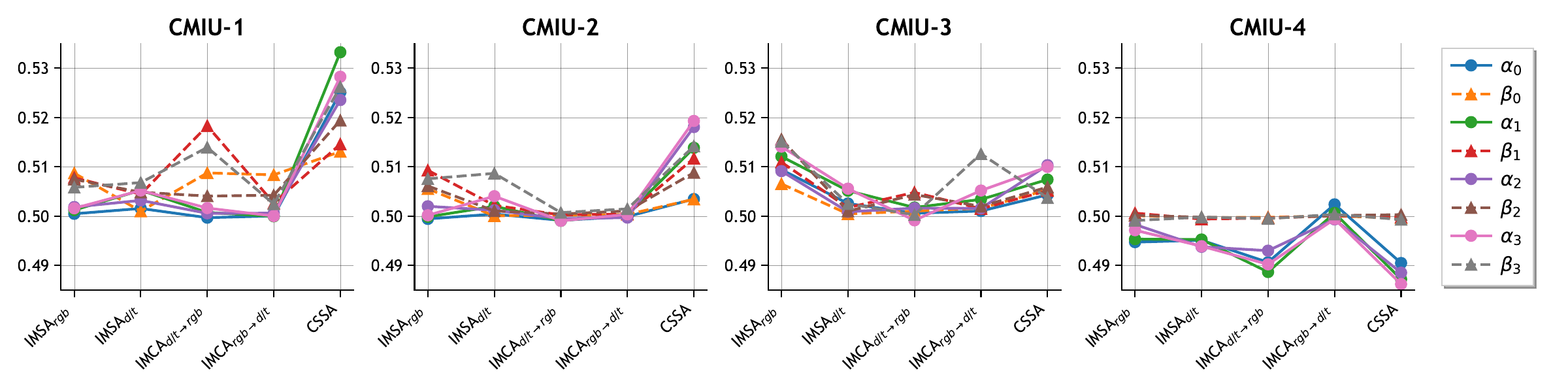}
  \caption{
    Visualization of the weights $\alpha$ and $\beta$ from different attention blocks in CMIU 1-4 of our four models: ``$*_{0}$'': ``Ours$^{50}$(I)'', ``$*_{1}$'': ``Ours$^{50}$(II)'', ``$*_{2}$'': ``Ours$^{101}$(I)'' and ``$*_{3}$'': ``Ours$^{101}$(II)''.
    ``IMSA$_x$'': The intra-modal self-attention block from the stream corresponding to the $x$ modality.
    ``IMCA$_{a \rightarrow b}$'': The inter-modal cross-attention block with the input $b$ for generating $Q$ and the input $a$ for generating $K$ and $V$.
    ``CSSA'': The cross-scale self-attention block.
  }
  \label{fig:weights}
\end{figure*}

\subsection{Ablation Analysis}\label{sec:ablation}

To evaluate the effectiveness of the proposed components and investigate their importance and contributions, based on the model ``Ours$^{50}$(I)'', we construct the ablation study on RGB-D SOD datasets.
At first, we build a CNN-based FPN-like two-stream model as the baseline network in which the features with the same resolution from both modalities are added directly.
As can be seen from Tab.~\ref{tab:ablation}, it is a sufficiently strong baseline model, which also makes the gains from our approach more reliable.
We also attach some other ablation analyses about the visualization of attention maps and some typical failure cases.

\paragraph{Effectiveness of Inter-Modal Cross-Attention.}
In our method, the inter-modal cross-attention enables the interaction and fusion of information between modalities by aligning related representations and gathering global context cues.
In the ablation analysis in Tab.~\ref{tab:ablation}, we show its performance improvement compared to the baseline model.
The significant improvement in average performance demonstrates the effectiveness of such a structural design.

\paragraph{Effectiveness of Inter-Modal and Cross-Scale Self-Attention.}
The intra-modal and cross-scale self-attention blocks are located at different locations in the information flow of the model and therefore act in different roles.
We show a careful ablation analysis in Tab.~\ref{tab:ablation} and Fig.~\ref{fig:visualablation} for different combinations of them.
The results in the table and the visual comparison reflect that they all have a positive effect on the model and the best performance is achieved when they coexist.
It further leads us to wonder whether a better performance could be obtained by repeatedly stacking more such structures.
This may be beyond the original intention of our structural design, but it is worth exploring further in the future.
Besides, we also try to replace these three self-attention blocks with the shifted window based self-attention block from the classic work of the vision transformer, Swin Transformer~\cite{Swin}.
In each block, a window-based self-attention module and a shifted one are placed sequentially.
The window size and the number of heads are set to 7 and 2.
Layer Normalization is used to normalize internal feature maps.
As can be seen from the comparison in Tab.~\ref{tab:cmp_more} and Tab.~\ref{tab:ablation},
with a similar amount of parameters and FLOPs, our method has higher FPS and better average performance, which indicates the competitiveness of the proposed structure.

\paragraph{Region-wise or Position-wise?}
The utilization of the local context is one of the key factors for the success of convolution in 2D vision tasks.
Although the transformer can facilitate the propagation of global information, it does not pay enough attention to the local region, which may lead to important local details being overlooked.
To alleviate this problem, we introduce locality to assist the attention operations.
The Conv-FFN (Sec.~\ref{par:cffn}) is used to replace the original position-wise FFN.
The comparison between Row 5 and Row 7 in Tab.~\ref{tab:ablation} reflects that the region-wise operation in the FFN affects the overall performance.

\begin{table*}[!t]
  \centering
  \caption{
    Ablation analysis of the different components.
    The best results are highlighted using {\color{reda} \textbf{red}}.
    When ``NUM=0'', the model is equivalent to the baseline.
  }
  \label{tab:ablation_cmiu}
  \resizebox{\linewidth}{!}{%
    \rowcolors{2}{gray!10}{white}
    \begin{tabular}{c*{4}{|*{5}{c}}}
	\toprule[2pt]
	                               & \multicolumn{5}{c|}{\textbf{NJUD}} & \multicolumn{5}{c|}{\textbf{SIP}} & \multicolumn{5}{c|}{\textbf{STEREO1000}} & \multicolumn{5}{c}{\textbf{AVERAGE}}                                                                                                                                                                                                                                                                                                                                                                                                         \\
	\multirow{-2}{*}{\textbf{NUM}} & $S_{m}\uparrow$                    & $F^{\omega}_{\beta}\uparrow$      & $M\downarrow$                            & $E_{m}\uparrow$                      & $F_{\beta}\uparrow$ & $S_{m}\uparrow$ & $F^{\omega}_{\beta}\uparrow$ & $M\downarrow$ & $E_{m}\uparrow$ & $F_{\beta}\uparrow$ & $S_{m}\uparrow$ & $F^{\omega}_{\beta}\uparrow$ & $M\downarrow$ & $E_{m}\uparrow$ & $F_{\beta}\uparrow$ & $S_{m}\uparrow$               & $F^{\omega}_{\beta}\uparrow$  & $M\downarrow$                 & $E_{m}\uparrow$               & $F_{\beta}\uparrow$           \\ \midrule[1pt]
	4                              & 0.921                              & 0.901                             & 0.031                                    & 0.953                                & 0.925               & 0.893           & 0.864                        & 0.042         & 0.933           & 0.902               & 0.913           & 0.882                        & 0.033         & 0.949           & 0.912               & {\color{reda} \textbf{0.909}} & {\color{reda} \textbf{0.882}} & {\color{reda} \textbf{0.035}} & {\color{reda} \textbf{0.945}} & {\color{reda} \textbf{0.913}} \\
	3                              & 0.918                              & 0.897                             & 0.033                                    & 0.950                                & 0.922               & 0.892           & 0.863                        & 0.043         & 0.933           & 0.901               & 0.909           & 0.874                        & 0.035         & 0.946           & 0.907               & 0.906                         & 0.878                         & 0.037                         & 0.943                         & 0.910                         \\
	2                              & 0.919                              & 0.896                             & 0.032                                    & 0.951                                & 0.925               & 0.889           & 0.856                        & 0.045         & 0.930           & 0.895               & 0.909           & 0.873                        & 0.035         & 0.945           & 0.907               & 0.906                         & 0.875                         & 0.037                         & 0.942                         & 0.909                         \\
	1                              & 0.919                              & 0.893                             & 0.033                                    & 0.945                                & 0.921               & 0.882           & 0.842                        & 0.051         & 0.918           & 0.885               & 0.909           & 0.873                        & 0.036         & 0.943           & 0.906               & 0.903                         & 0.869                         & 0.040                         & 0.935                         & 0.904                         \\
	0                              & 0.917                              & 0.891                             & 0.034                                    & 0.939                                & 0.922               & 0.883           & 0.847                        & 0.048         & 0.924           & 0.891               & 0.903           & 0.862                        & 0.039         & 0.935           & 0.899               & 0.901                         & 0.867                         & 0.040                         & 0.933                         & 0.904                         \\
	\bottomrule[2pt]
\end{tabular}

  }
\end{table*}

\begin{table*}[!t]
  \centering
  \caption{
    Ablation analysis of the path size for the PTRE in different decoding levels.
    In these forms, the four numbers correspond to the four levels from top to bottom, respectively.
    ``\blank'': Out of memory.
  }
  \label{tab:patchsize}
  \resizebox{\linewidth}{!}{%
    \rowcolors{2}{gray!10}{white}
    \begin{tabular}{c|c*{4}{|*{5}{c}}}
	\toprule[2pt]
	                                       & \textbf{FLOPs} & \multicolumn{5}{c|}{\textbf{NJUD}} & \multicolumn{5}{c|}{\textbf{SIP}} & \multicolumn{5}{c|}{\textbf{STEREO1000}} & \multicolumn{5}{c}{\textbf{AVERAGE}}                                                                                                                                                                                                                                                                                                                                                                                                             \\
	\multirow{-2}{*}{\textbf{PATCH SIZE }} & \textbf{(G)}   & $S_{m}\uparrow$                    & $F^{\omega}_{\beta}\uparrow$      & $M\downarrow$                          & $E_{m}\uparrow$                      & $F_{\beta}\uparrow$ & $S_{m}\uparrow$ & $F^{\omega}_{\beta}\uparrow$ & $M\downarrow$ & $E_{m}\uparrow$ & $F_{\beta}\uparrow$ & $S_{m}\uparrow$ & $F^{\omega}_{\beta}\uparrow$ & $M\downarrow$ & $E_{m}\uparrow$ & $F_{\beta}\uparrow$ & $S_{m}\uparrow$               & $F^{\omega}_{\beta}\uparrow$  & $M\downarrow$               & $E_{m}\uparrow$               & $F_{\beta}\uparrow$           \\ \midrule[1pt]
	$1,1,1,1$                              & 71.853         & \blank                             & \blank                            & \blank                                   & \blank                               & \blank              & \blank          & \blank                       & \blank          & \blank          & \blank              & \blank          & \blank                       & \blank          & \blank          & \blank              & \blank                        & \blank                        & \blank                        & \blank                        & \blank                        \\
	$2,2,2,2$                              & 50.907         & 0.918                              & 0.896                             & 0.032                                    & 0.948                                & 0.920               & 0.889           & 0.860                        & 0.044           & 0.931           & 0.898               & 0.914           & 0.882                        & 0.033           & 0.949           & 0.911               & 0.907                         & 0.879                         & 0.036                         & 0.943                         & 0.910                         \\
	$4,4,4,4$                              & 45.731         & 0.917                              & 0.896                             & 0.033                                    & 0.948                                & 0.920               & 0.893           & 0.865                        & 0.042           & 0.934           & 0.903               & 0.912           & 0.880                        & 0.034           & 0.948           & 0.909               & 0.907                         & 0.880                         & 0.036                         & 0.943                         & 0.911                         \\
	$\mathbf{8,8,8,8}$                     & 44.442         & 0.921                              & 0.901                             & 0.031                                    & 0.953                                & 0.925               & 0.893           & 0.864                        & 0.042           & 0.933           & 0.902               & 0.913           & 0.882                        & 0.033           & 0.949           & 0.912               & {\color{reda} \textbf{0.909}} & {\color{reda} \textbf{0.882}} & {\color{reda} \textbf{0.035}} & {\color{reda} \textbf{0.945}} & {\color{reda} \textbf{0.913}} \\
	$8,16,16,16$                           & 44.120         & 0.920                              & 0.900                             & 0.032                                    & 0.951                                & 0.924               & 0.892           & 0.863                        & 0.044           & 0.932           & 0.901               & 0.912           & 0.878                        & 0.034           & 0.947           & 0.908               & 0.908                         & 0.880                         & 0.037                         & 0.943                         & 0.911                         \\
	$8,16,32,32$                           & 44.039         & 0.915                              & 0.894                             & 0.033                                    & 0.948                                & 0.917               & 0.883           & 0.851                        & 0.047           & 0.923           & 0.891               & 0.914           & 0.884                        & 0.033           & 0.950           & 0.912               & 0.904                         & 0.876                         & 0.038                         & 0.940                         & 0.907                         \\
	$8,16,32,64$                           & 44.021         & 0.916                              & 0.893                             & 0.034                                    & 0.947                                & 0.917               & 0.890           & 0.861                        & 0.042           & 0.933           & 0.898               & 0.911           & 0.878                        & 0.035           & 0.947           & 0.908               & 0.906                         & 0.877                         & 0.037                         & 0.942                         & 0.908                         \\
	\bottomrule[2pt]
\end{tabular}

  }
\end{table*}

\paragraph{Analysis of View-Mixed Attention.}
The good performance of the proposed view-mixed attention can be seen from the aforementioned experiments and comparisons.
To further verify the effectiveness of this multi-view attention variant, in Tab.~\ref{tab:casaattn_ffn}, we compare the performance of the two branches using different combination strategies based on the model ``Ours$^{50}$(I)'' and our strategy shows the best performance.
It is worth noting that when both weights are fixed at $0.5$, the model is also competitive.
Besides, from the summary of the weights in different CMIUs in Fig.~\ref{fig:weights}, the learned $\alpha$ and $\beta$ are around 0.5.
The differences between learned parameters in different positions further reflect the flexibility of our method, which also enables our method to remain optimal in most metrics.
Besides, the figure also reflects some interesting phenomena.
In the shallow layer, the overall weight values are relatively larger, and the channel branch in the cross-modal interaction block IMCA plays a more important role.
This also implies the necessity of introducing shallow features as well as the channel view.
Besides, despite the differences in model capacity and training data, the trends in the four model weights are very similar.
The underlying reason is still worth exploring, and it also further highlights the potential of the channel-wise interaction pattern in the feature decoding stage.

\begin{figure}[!t]
  \centering
  \includegraphics[width=0.9\linewidth]{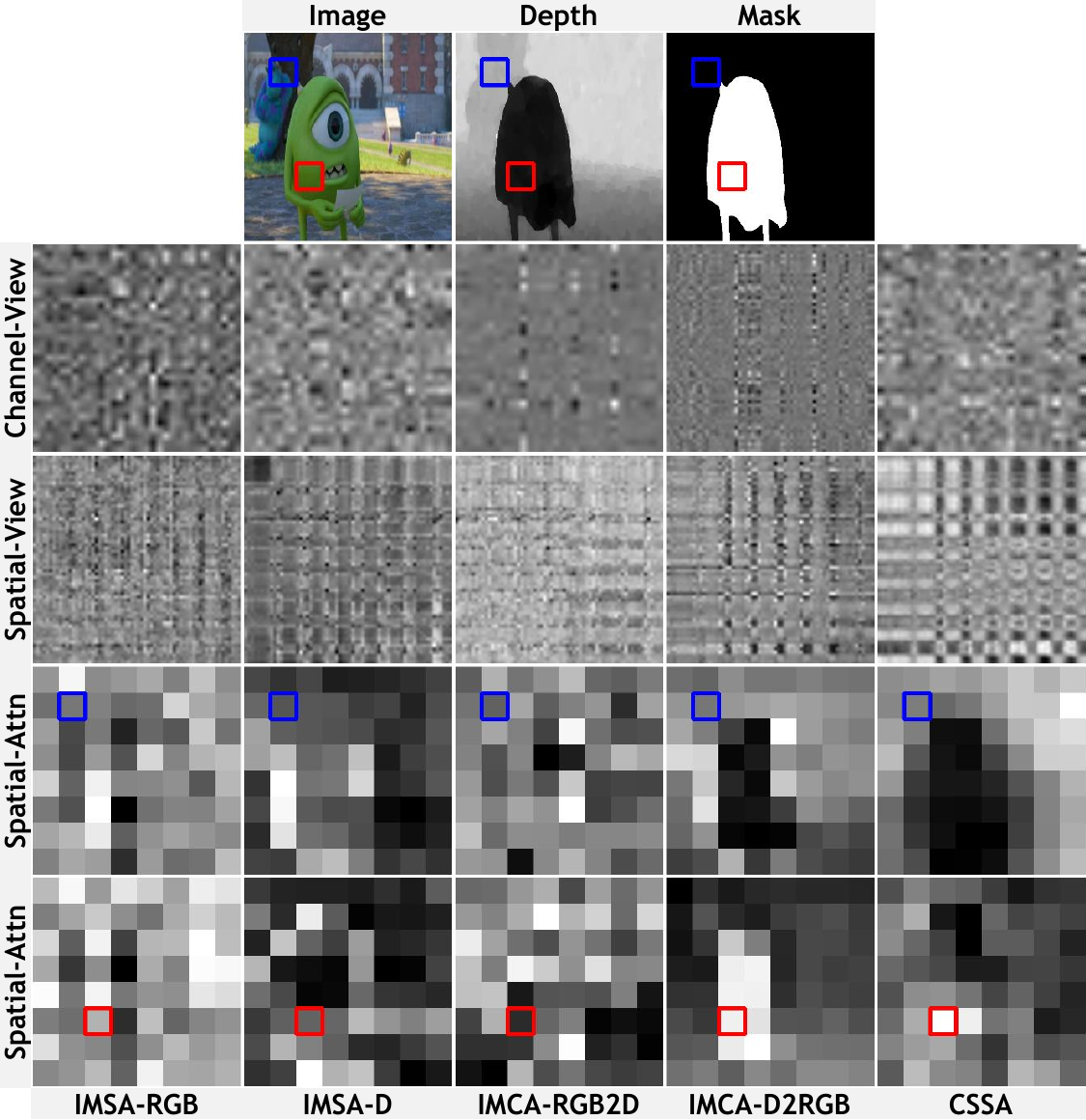}
  \caption{
    Visualization of different attention operations in the $1^{st}$ CMIU.
    In these maps, normalization and upsampling based on the bilinear interpolation are applied to show them more clearly.
    Besides, we also visually compare the spatial attention maps of two different queries marked by the {\color{blue} \textbf{blue}} and {\color{red} \textbf{red}} squares from the background and foreground of the image.
    ``Channe-View'': The pairwise channel similarity score map from the channel-view attention operation.
    ``Spatial-View'': The pairwise spatial similarity score map from the patch-wise spatial-view attention operation.
    ``Spatial-Attn'': The global spatial attention map corresponding to the position of the {\color{blue} \textbf{blue}} and {\color{red} \textbf{red}} squares.
    ``IMSA-$x$'': The maps from the IMSA corresponding to the $x$ modality.
    ``IMCA-$a$2$b$'': The maps from the IMCA with the input $b$ for generating $Q$ and the input $a$ for generating $K$ and $V$.
    ``CSSA'': The maps from the CSSA.
  }
  \label{fig:attnmap}
\end{figure}

\paragraph{Number of Stages with the CMIU.}
To explore the effect of the CMIU, we gradually replace the CMIU in the original CAVER with the simple convolutional layer in the order from shallow to deep.
The experimental results are listed in Tab.~\ref{tab:ablation_cmiu}.
It can be seen from the table that with the gradual removal of the CMIU from shallow to deep, the performance of the model has been consistently decreased on multiple datasets.

\begin{figure*}[!t]
  \centering
  \includegraphics[width=\linewidth]{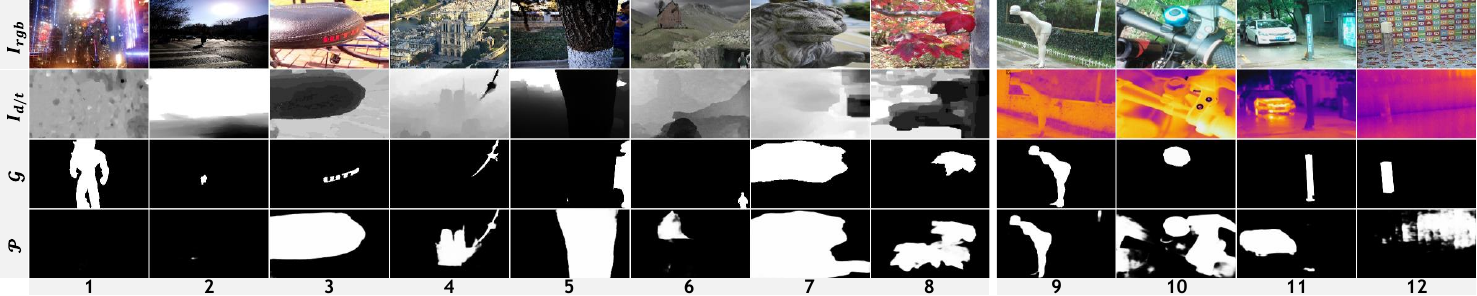}
  \caption{
    Visualization of some typical failure cases from RGB-D and RGB-T test datasets.
  }
  \label{fig:failcase}
\end{figure*}

\paragraph{Patch Size of the PTRE.}
The size of the patch re-embedding in the PTRE is an important hyperparameter.
We compare several settings in Tab.~\ref{tab:patchsize}.
It should be noted that ``$1,1,1,1$'' is actually equivalent to the standard attention operation~\cite{Transformer}, which cannot be trained on our device due to its large memory requirements.
As shown in the table, compared with other settings, ``$8,8,8,8$'' has the best average performance and a moderate computational cost, so, it is used as the default setting in our other experiments.
This also reflects that simply increasing the computational cost does not necessarily lead to a better performance, which sometimes depends on a more reasonable structural design.
An interesting phenomenon occurs under the setting of the final line of Tab.~\ref{tab:patchsize}.
In our experiments, the input image size is 256, and the sizes of the four feature maps extracted from the encoder are 64, 32, 16, and 8, respectively.
So in this case, the size of the attention maps in all spatial branches is $1 \times 1$, and the process of gathering information from the value $V$ is simplified to scale $V$ using a specific scalar factor.
Its performance is not as good as the setting ``$8,8,8,8$'', but it is still competitive.
The existence of the channel-view component does play an important role in such competitive performance.
It builds dense interactions between feature channels by performing dynamic feature fusion of different global views of the entire image.
So, it can provide effective assistance for such a simple model.
Considering its more efficient computation, it can be the basis for exploring lighter model variants.

\paragraph{Visualization of Attention Maps.}
To visualize the impact of attention operations at different positions, we show in Fig.~\ref{fig:attnmap} the learned pair-wise similarity score map which is calculated by the dot product between the patch-wise query and key.
To show the learning effect more intuitively, we also supplement the normalized alignment score map of the query token corresponding to the red marker in RGB and depth images and the global key token sequence.
It can be noticed that in the whole process, the feature discriminativeness and the intra-class similarity are significantly enhanced.
From another standpoint, the global dependency pattern can be clearly observed in each map of different blocks, which is the source of the power of the attention operation.

\paragraph{Analysis of Typical Failure Cases.}\label{par:failcase}
We show some typical failure cases in Fig.~\ref{fig:failcase}.
The visualization results mainly involve the following three very thorny problems:
\begin{enumerate}
  \item \textbf{Uncertainty} of the object. In Case 1, 2, 10, and 12, the object of interest is difficult to be completely captured due to the extremely complex scene. The location and segmentation of such objects rely on the model's ability to adapt to such significant environmental interference. Our architectural form based on global relationship modeling may lead to weaker performance for such objects.
  \item \textbf{Ambiguity} of the definition. In our collection, the definition of saliency objects in some samples (Case 3-6, 8, and 11 in Fig.~\ref{fig:failcase}) do not meet the criteria for visual saliency. The performance loss brought by these samples is difficult to ignore for some bi-modal SOD datasets with small data volumes involved in this paper. But our model does segment those visually more salient objects well. This can actually provide some meaningful references for future work on the label calibration task.
  \item \textbf{Misleadingness} of the annotation image. As shown in cases 7 and 9 in Fig.~\ref{fig:failcase}, our method yields visually more accurate predictions for these RGB images, but incomplete or unaligned masks from the test dataset cause unexpected errors in the evaluation process.
\end{enumerate}
The aforementioned issues are still very challenging now, and these will be the focus of our future work.

\section{Conclusion}

In this paper, we rethink the architecture design for the bi-modal SOD task.
A novel view-mixed transformer-based top-down information propagation path is proposed to enhance the capability of perceiving and excavating the important global cues in intra-modal features and simplify the cross-modal long-range interaction and alignment.
And by using the PTRE, the computational and storage intensity of the matrix operation in the attention block is effectively reduced, which makes it possible to process multi-scale high-resolution features.
In addition, the convolutional FFN further enhances the critical local details in the feature map.
Extensive experimental comparisons show the effectiveness of the proposed method.

\bibliographystyle{IEEEtran}
\bibliography{IEEEabrv,egbib.bib}

\end{document}